\documentclass[10pt,twocolumn,letterpaper]{article}

\usepackage[pagenumbers]{cvpr} 

\usepackage{amsthm}
\usepackage{newtxtext}
\usepackage[varg]{newtxmath}
\usepackage{amssymb}
\usepackage{url}
\usepackage{float}
\usepackage{booktabs}
\usepackage[dvipsnames]{xcolor}

\usepackage{algorithm}
\usepackage[noend]{algpseudocode} 
\usepackage{algorithmicx}

\usepackage{bm}
\definecolor{OursBlue}{RGB}{224,242,255}
\usepackage[table,dvipsnames]{xcolor}

\definecolor{cvprblue}{rgb}{0.21,0.49,0.74}
\usepackage[pagebackref,breaklinks,colorlinks,allcolors=cvprblue]{hyperref}

\def\paperID{*****} 
\def\confName{CVPR}
\def\confYear{2026}

\title{\textit{OrnaStyler}: Ornament-Aware Latent Editing for \\ Content-Preserving 3D Stylization}

\author{
Tomohiro Aizawa$^{1}$\qquad
Shigeru Kuriyama$^{2,3}$ \qquad
Chunzhi Gu$^{1*}$ 
\\
$^{1}$University of Fukui, Japan\\
$^{2}$Toyohashi University of Technology, Japan\\
$^{3}$AI Lab, CyberAgent, Inc., Japan
}

\begin{document}

\twocolumn[{
\renewcommand\twocolumn[1][]{#1}
\maketitle
\begin{center}
    \centering
    \captionsetup{type=figure}
    \includegraphics[width=0.99\linewidth]{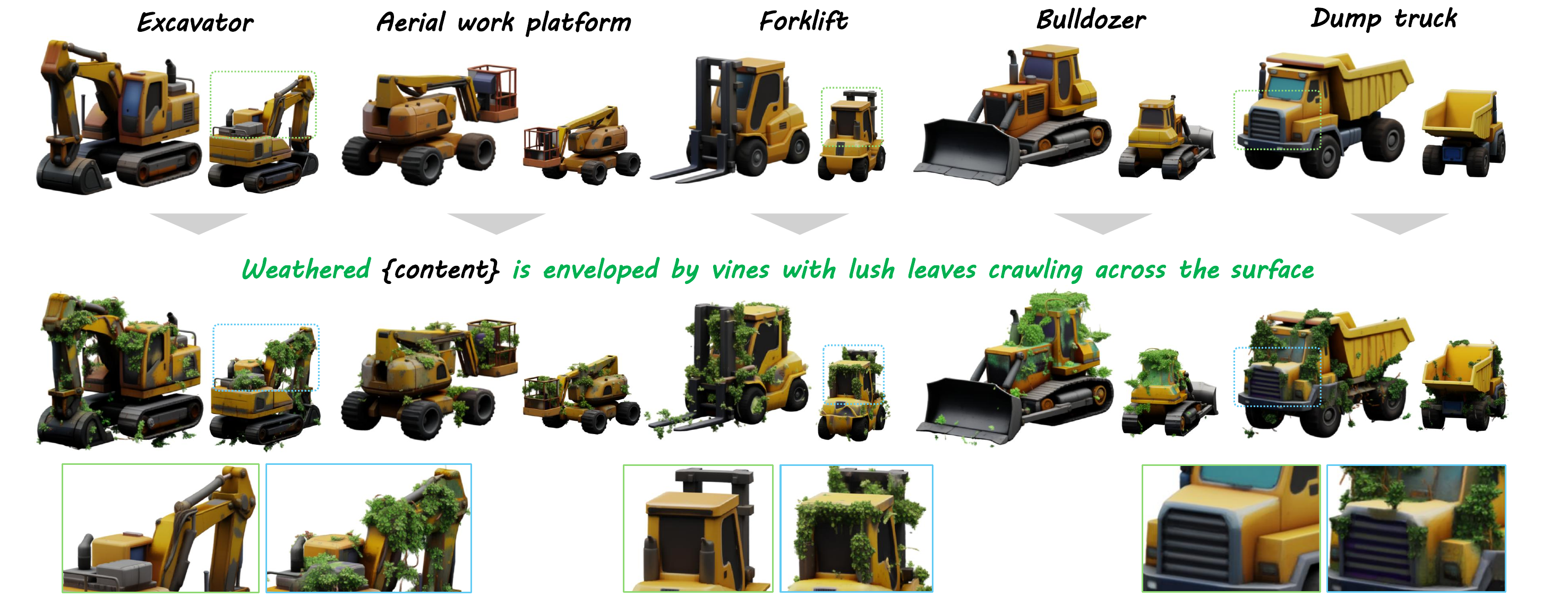}
    \caption{\textbf{OrnaStyler} enables consistent, zero-shot, text-guided 3D stylization from the input content assets (top row) to stylized results (bottom row). Our method specifically models the ornament structures to seamlessly integrate with the underlying texture and geometry, which jointly pursues \textit{\textbf{content}} preservation and \textit{\textbf{\textcolor{Green}{style}}} reflection.}
    \label{fig:teaser}
\end{center}
}]

\begingroup
\renewcommand{\thefootnote}{\fnsymbol{footnote}}
\footnotetext[1]{Corresponding Author.}
\endgroup

\begin{abstract}
 Text-guided style editing of 3D assets is essential for adapting existing objects to diverse visual aesthetics in digital content creation. Despite rapid progress in 3D shape modeling, faithfully stylizing an existing asset remains challenging when the desired stylization involves fine-grained structural ornamentation, which requires the model to preserve the source geometry and object identity, while coherently integrating new style-specific details. We propose \textbf{OrnaStyler}, a zero-shot framework for text-guided ornament-aware 3D stylization. Built upon rectified flow-based generative modeling, OrnaStyler introduces an inversion-guided editing strategy that recovers content-aware latent representations at both geometry and appearance levels in a staged manner to facilitate faithful editing. Our core idea is to explicitly model the spatial configuration of stylistic elements, thereby mitigating the fundamental tension between content preservation and style expression in the voxel space. Specifically, at the geometry level, we manipulate voxel representations through flow inversion to synthesize ornament-enhanced structures while preserving the spatial identity of the source asset. Then, at the appearance level, we introduce an adjacency-aware feature inpainting mechanism to harmonize newly generated ornaments with the original content, yielding coherent geometry-appearance integration. Our approach operates solely in the inference phase and enables selective editing over geometric augmentation or appearance stylization. Extensive experiments on both generated and real-world 3D assets against prior methods demonstrate that OrnaStyler achieves state-of-the-art editing performance in terms of content preservation, style fidelity, and overall visual realism. 
 Code is available at: \url{https://github.com/tomohiro0427/OrnaStyler}. 
\end{abstract}  


\section{Introduction}
Recent advancements in 3D asset generation models have enabled the creation of high-quality 3D assets from diverse types of prompts, such as texts \cite{poole2023dreamfusion,tang2024lgm,xiang2025structured}, or even a single image \cite{qian2024magic123,liu2023zero,tang2023make, hunyuan3d22025tencent}. Beyond creating new assets from scratch, practical 3D workflows also require style-specific editing, where an existing asset is adapted to a desired artistic style while retaining its recognizable content identity. Practically, style editing (i.e., stylization) techniques for 3D assets are essential for scalable digital content creation across various industries, including gaming, AR/VR, and filmmaking.

\begin{figure}[t]
    \centering
    \includegraphics[width=\linewidth]{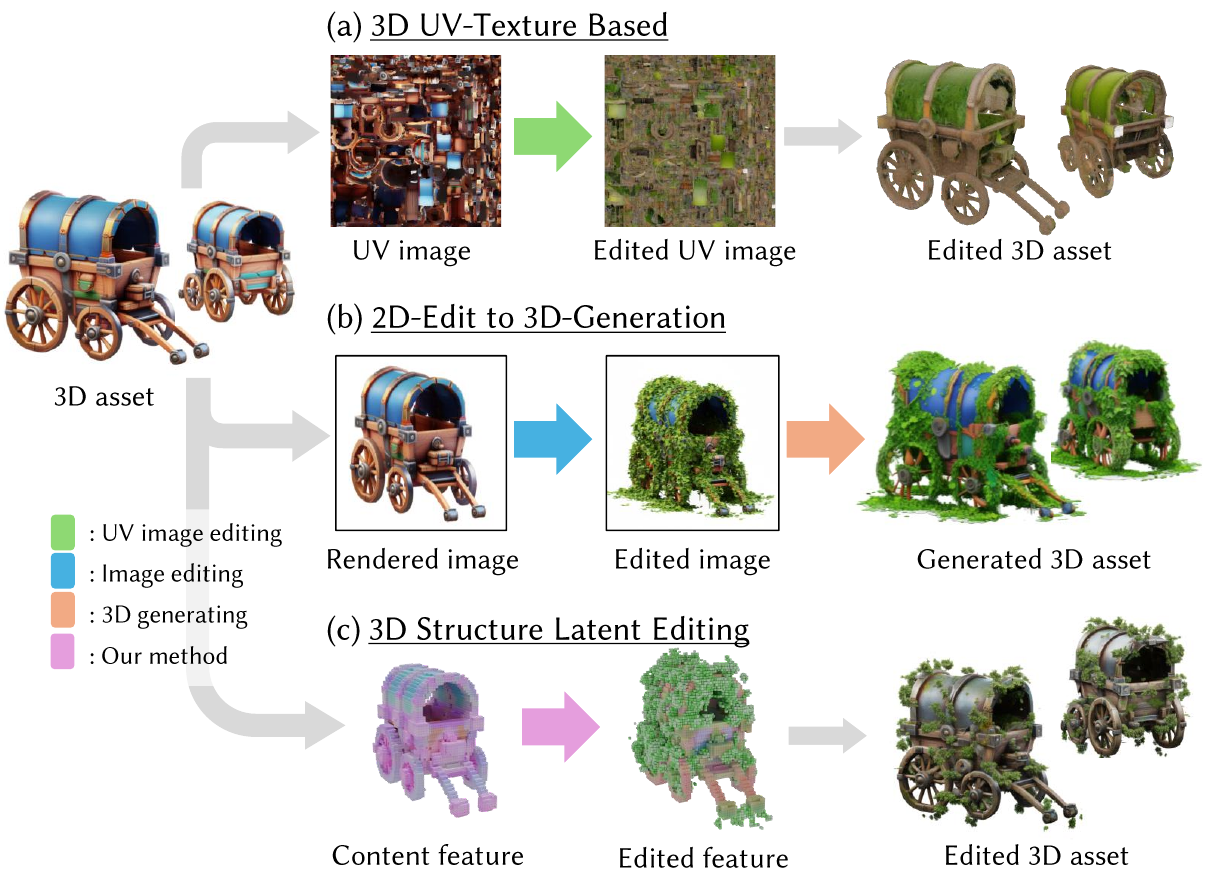}
    \caption{\textbf{Conceptual comparison} of 3D style editing approaches. In contrast to \textit{UV Texture-based} (a) or \textit{2D-Edit to 3D-Generation} (b) paradigms, our \textit{3D structure latent-guided} (c) paradigm yields the most natural and content-preserving editing effects.}
    \label{fig: intro}
\end{figure}

In addition to style fidelity, one key challenge in 3D stylization lies in introducing style-specific \textbf{structural ornamentation} while preserving the source geometry and object identity, which remains underexplored to date. In particular, these techniques are mostly text-based, and can be categorized into two paradigms: (i) \textbf{UV-texture-based methods} \cite{richardson2023texture, zeng2024paint3d, liu2024text} extract or reconstruct the UV texture map from the content geometry, apply text-driven stylization in the 2D texture domain, and remap the result back to the 3D asset. Although such approaches can reflect style cues (Fig.~\ref{fig: intro}(a)), the inherent domain gap between 2D and 3D often triggers inconsistent appearance and geometric artifacts; (ii) \textbf{2D-edit-3D-generation methods} \cite{comanici2025gemini, zhang2025enabling, xiang2025structured} edit rendered images and reconstruct a 3D asset from the edited views. While the edited images are visually stylized, the 3D reconstruction phase typically disregards the original geometric constraints, resulting in significant structural deviations from the content asset (Fig.~\ref{fig: intro}(b)). More importantly, neither paradigm explicitly models the spatial configuration of the required ornamentation during 3D stylization, and ornament structures are either reduced to 2D appearance changes or indirectly hallucinated through reconstruction, often causing implausible geometry between content and ornaments.

In this work, we ask: \textit{instead of relying on 2D intermediates to transfer style and then reconstruct geometry, is it possible to introduce style-specific ornamentation directly in 3D space while keeping the source asset intact?} Our core insight is that ornament-aware stylization should be viewed not merely as appearance transfer, but as the carefully designed spatial organization of additional structures around the existing content. Yet, since the desired ornament structures are not explicitly given and must remain compatible with the source geometry and appearance, how to localize and harmonize them remains challenging.

To this end, we propose \textbf{OrnaStyler}, a novel text-guided style editing framework for 3D assets that explicitly targets ornament-aware stylization. OrnaStyler is built upon a pre-trained rectified flow (RF)-based 3D asset generation pipeline \cite{xiang2025structured} and follows a two-stage modeling paradigm to progressively operate at the voxel and appearance levels during generation.
Instead of dealing with ornamentation as a purely texture-centric effect, OrnaStyler first determines the spatial configurations (i.e., voxel coordinates) of ornamented structures to reflect style cues at the geometric level. Specifically, we incorporate a flow inversion mechanism to trace back the initial seed latent corresponding to the content voxel representation, and then regenerate ornament-enhanced voxels conditioned on the style prompts using the RF generation process. Since the inverted latent compactly encodes structural information of the original content, manipulating it allows us to adaptively localize the ornament geometries while preventing undesired corruption of the content shape (Fig.~\ref{fig: intro}(c)). Importantly, OrnaStyler mitigates the inherent spatial competition between content and stylistic structures by explicitly allocating exclusive active voxels for ornaments, which enables style elements to be introduced without compromising the integrity of the original geometry.

The second challenge is to generate the appearance-level latent feature embeddings for our structured voxels. To address this, we propose to employ a 3D inpainting strategy to predict the features for the ornaments based on the known content features for visually coherent and harmonized stylization. However, we observe that naïve inpainting may still introduce occasional distortions near the boundaries between content and newly added ornament regions. To bypass this barrier, we incorporate a spatial mask into the inpainting process that explicitly captures adjacency between ornaments and content to further promote boundary-aware feature integration. Similar to the voxel-level editing stage, we again employ flow inversion to identify seed feature latents that faithfully anchor the editing process to the original content representation. The resulting voxel- and feature-level representations are eventually fed into a pre-trained object decoder to synthesize the edited 3D asset.

As OrnaStyler is training-free, it serves as a convenient yet efficient zero-shot text-guided framework for 3D assets to enable flexible stylization that targets ornament-aware structural editing. Extensive experiments across diverse forms of objects and style prompts demonstrate state-of-the-art editing effectiveness against prior competitors, regarding realism, content consistency, and style fidelity. 
Our contributions can be summarized as follows: 
\begin{itemize}
\item 
Identifying the inherent limitations of existing 3D stylization paradigms in balancing stylistic elements (i.e., ornaments) and content geometry within constrained spatial capacity.
\item 
Proposing a novel zero-shot framework, \textbf{OrnaStyler}, which explicitly models spatial configurations for ornament-enhanced geometries and leverages an inversion-based mechanism to derive seed latents for content-preserving stylization.
\item 
Developing a 3D inpainting-inspired feature generation strategy that enables seamless appearance harmonization between style and content via adjacency-aware masking for boundary-consistent synthesis.
\end{itemize}

\section{Related work}
\subsection{3D Generative Models } \label{sec: related_generative_model}
Early attempts at 3D generation focus on producing geometric shapes using various representations, such as point clouds \cite{zhou20213d,mo2023dit}, meshes \cite{zeng2022lion, Liu2023MeshDiffusion}, and Signed Distance Fields (SDFs) \cite{chou2023diffusion,yariv2024mosaic}, typically leveraging diffusion models \cite{luo2021diffusion, zheng2023lasdiffusion} or Generative Adversarial Networks (GANs) \cite{chan2022efficient, zheng2022sdfstylegan}.
Later approaches \cite{jain2022zero, lin2023magic3d, tang2023make} have aimed to extend the advantages of 2D diffusion models to 3D generation by distilling 3D information with pretrained 2D diffusion models via Score Distillation Sampling (SDS) \cite{poole2023dreamfusion}. Despite encouraging results, these methods often suffer from limited fidelity and inefficient optimization. To address these efficiency concerns, several subsequent works \cite{xiong2025octfusion, deng2025efficient, wei2025octgpt} have explored different generation processes with more compact latent representations.
Recently, large-scale 3D generative models \cite{xiang2025structured, hunyuan3d22025tencent, li2025triposg, xiang2026native} have demonstrated remarkable performance by training on extensive datasets, enabling the production of high-quality 3D assets through flow-based generation. A pioneering work is TRELLIS \cite{xiang2025structured}, which learns a highly structured latent representation that can be decoded into the texture and geometric layout of a 3D asset. Its extension, TRELLIS.2 \cite{xiang2026native}, further supports image-conditioned generation using field-free sparse voxel representations. As we target text-guided stylization, we utilize TRELLIS as our base model to leverage the robust capability in generating diverse and high-fidelity 3D designs.

\subsection{3D Style Editing} \label{sec: related_style_editing}
\noindent \textbf{Image-Guided Approaches.}
Given a reference style image, 3D style transfer aims to synthesize a 3D asset that reflects the visual style of the reference while preserving the original content.
Early methods \cite{fan2022unified, zhang2022arf, liu2023stylerf, pang2023locally, zeng2024ipdreamer} primarily perform stylization in NeRF-based representations by directly optimizing color and appearance to match the reference image. For example, IPdreamer \cite{zeng2024ipdreamer} introduced image-prompt score distillation sampling to capture fine-grained appearance cues from complex visual inputs. However, the NeRF-based techniques  \cite{fan2022unified, zhang2022arf, liu2023stylerf} typically rely on iterative scene-wise optimization, resulting in substantial computational overhead and limited scalability. To address this limitation, more recent methods \cite{kim2024fprf, oztas20253d} have shifted to leverage large-scale pre-trained 3D reconstruction or generative models to enable feed-forward stylization for efficiency. Another line of work focuses on texture-based stylization \cite{xie2024styletex,yeh2024texturedreamer,zeng2024paint3d}, where diffusion models are used to generate stylized textures for 3D meshes. While these approaches can produce visually appealing results, they often fail to preserve the identity of the original content, including both geometry and appearance. To address the issues of content leakage and style deviation during transfer, several works \cite{chen2024scenetex, xie2024styletex, qu2025stylesculptor} attempt to disentangle style and content representations to generate textures or geometries that exhibit the desired style.
Building upon this idea, more recent works \cite{qu2025stylesculptor,liu2026interpd, sun2026morphany3d} leveraged large-scale 3D generative models to synthesize stylized 3D assets conditioned on both content and reference-style images in a zero-shot manner to enable the integration of content and style information. However, faithfully preserving the original content remains challenging, as reference style images often contain complex structures that can interfere with the underlying geometry and tend to induce content inconsistencies.

\begin{figure*}
    \centering
    \includegraphics[width=0.92\linewidth]{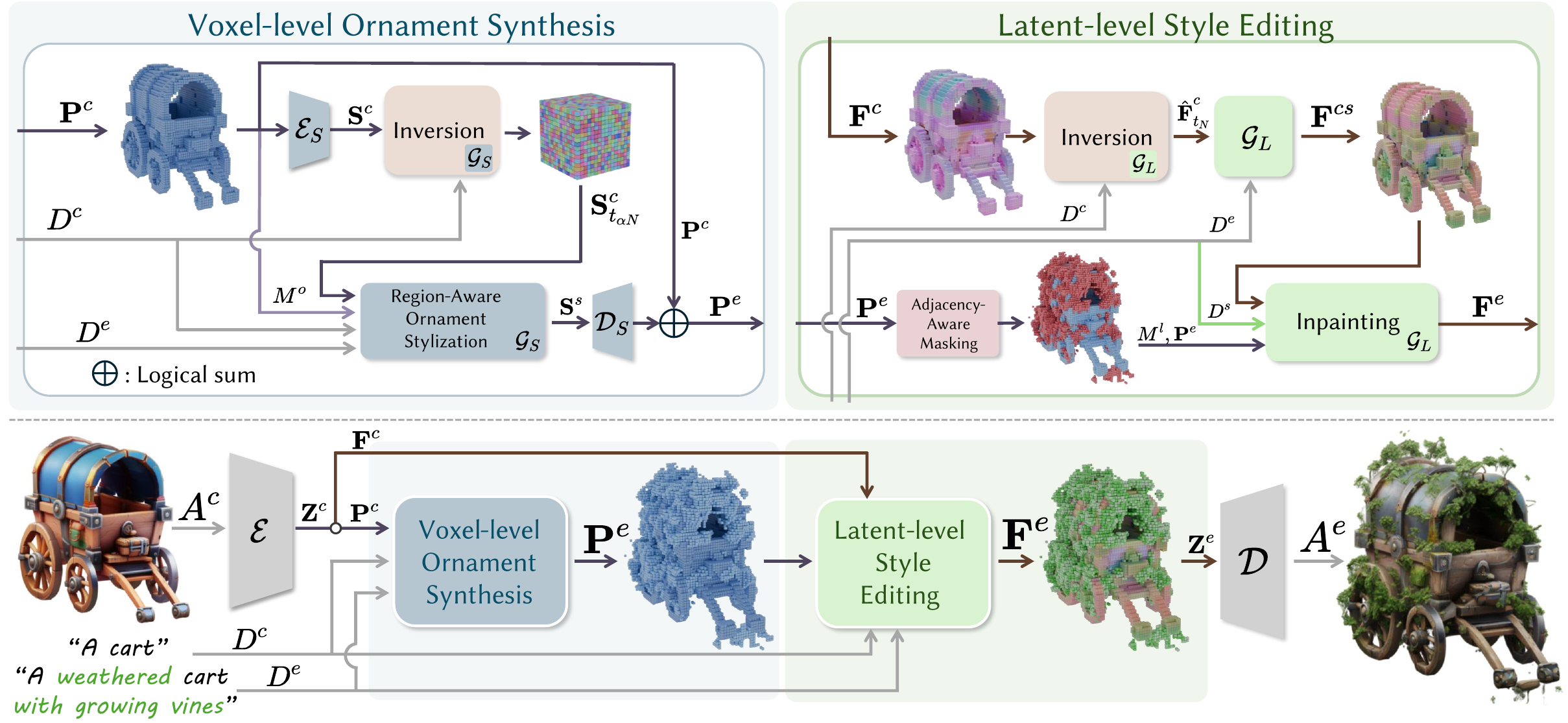}
    \caption{\textbf{Method overview of our proposed \textit{OrnaStyler}.}
Given a content asset $A^c$ produced from a content prompt $D^c$ and an edited prompt $D^e$, OrnaStyler generates a stylized 3D asset $A^e$ in a two-stage manner. 
It first generates in the \textbf{\textcolor[HTML]{2093BD}{Voxel-level ornament synthesis stage}} the ornament-enhanced voxel coordinates, and then produces the corresponding latent features 
in the \textbf{\textcolor[HTML]{54B331}{Latent-level style editing stage}} for appearance stylization, to eventually create the edited asset.  }
    \label{fig: pipeline}
\end{figure*}

\noindent \textbf{Text-Guided Approaches.}
Compared to image-based guidance, text-based conditioning provides a more concise and flexible interface for controlling target appearance. Most text-driven methods focus on stylizing UV texture maps to match textual descriptions. Specifically, early works \cite{richardson2023texture,chen2023text2tex,zhang2024texpainter} reconstruct UV textures by projecting multi-view generated images onto meshes, with additional refinements such as camera pose optimization \cite{chen2023text2tex} and trimap-based blending \cite{richardson2023texture}. However, these multi-view pipelines are computationally expensive due to per-view generation and often suffer from cross-view inconsistency. Some later attempts \cite{zeng2024paint3d,liu2024text} leveraged pre-trained text-to-image diffusion models to address these issues. For example, Zeng et al.~\cite{zeng2024paint3d} proposed a coarse-to-fine pipeline that first generates a coarse UV texture from two views and then refines details via inpainting. To further mitigate seam artifacts introduced by inpainting, Liu et al.~\cite{liu2024text} developed a synchronous multi-view diffusion framework that enforces cross-view consistency during denoising. Despite their effectiveness in appearance editing, these texture-centric methods are inherently limited in their ability to modify underlying geometry. More recent work \cite{xiang2025structured} addresses this limitation by leveraging 3D generative priors to enable appearance editing alongside moderate geometric adaptation while preserving coarse structure. However, faithfully maintaining fine-grained attributes, such as detailed color patterns and local designs, is still challenging. As our approach does not depend on UV texture maps and explicitly models both geometry and appearance during stylization, it enables content-preserving editing while introducing coherent style-specific structures.


\section{Preliminaries}
In this section, we give a brief introduction to the preliminaries of TRELLIS \cite{xiang2025structured}, which serves as our backbone 3D generation model.
Given the prompt $C$ provided in image or text, TRELLIS produces a Structured Latent (SLat), which represents the geometry and appearance clues in a unified manner, to eventually generate the 3D asset $A$. In particular, the SLat $\mathbf{Z}$ comprises two discrete components $\mathbf{Z} = \{ \mathbf{P}, \mathbf{F}\}$: the structural geometry $\mathbf{P} =\{ p_{j}\}^{L}_{j = 1}$ representing voxel coordinates, and the latent features $\mathbf{F} =\{ f_{j}\}^{L}_{j = 1}$ derived from aggregated multi-view features using DINOv2  \cite{oquab2024dinov2}, with a total of $L$ occupied voxels. 
Here, $p_{j} \in \{ 0,1, \dots, K-1\}^3$ indexes the position within a voxel grid of resolution $K$, and $f_{j} \in \mathbb{R}^{d}$ represents the latent feature corresponding to an active voxel.

\noindent \textbf{Rectified Flow.}
TRELLIS adopts the rectified flow (RF) model \cite{lipman2023flow} as its generative backbone to learn the latent pair ($\mathbf{P}, \mathbf{F}$). In particular, RF defines a forward process that progressively perturbs a data sample $x$ via linear interpolation $\psi$ between itself and a pure Gaussian noise $\bm{\epsilon} \sim \mathcal{N}(\mathbf{0}, \mathbf{I})$, following $\psi_{t_i}(x)=(1-t_i)x+t_i\bm{\epsilon}$, where $t_i \in [0,1] (t_i < t_{i+1})$ is an arbitrary timestep. Then, in the reverse process, RF learns a time-dependent velocity field $\bm{v}_{t_i}$ that transports the noisy observation back toward the clean data distribution, which is parameterized via a neural network $ \bm{v}(x_{t_i}, t_i, C)$ at the $t_i$-th timestep $ \bm{v}_{t_i} =\bm{v}(x_{t_i}, t_i, C)$. Following Euler's method, the reverse dynamics are discretized as 
\begin{equation}
x_{t_{i-1}} = x_{t_i} + (t_{i-1} - t_i)\bm{v}(x_{t_i},t_i,C),
\label{eq:flow_euler_solver}
\end{equation}
where $i \in \{ 0, \dots ,N\} $,  $ t_0 = 0$, $t_N = 1$, and $x_{t_0} = x$.
$N$ refers to the total number of denoising iteration steps. The learning process is achieved by \textit{minimizing} the conditional flow matching (CFM) loss \cite{lipman2023flow}: $\mathcal{L}=\mathbb{E}_{t_i,x,\bm{\epsilon}}|| \bm{v}(\psi_{t_i}(x),t_i,C) - (\bm{\epsilon} - x)||_2^2$.

\noindent \textbf{Asset Generation.}
Given the condition prompt $C$, TRELLIS generates the SLat $\mathbf{Z}$ in a staged manner to eventually produce the 3D asset $A$. This is achieved by preparing two individual rectified flow models $(\mathcal{G}_{S}, \mathcal{G}_{L})$, serving as the sparse structure generator and the structured latent generator, respectively. 
In the first stage, $\mathcal{G}_{S}$ is employed to produce sparse structural information by sampling an initial noisy voxel latent $\mathbf{S}_{t_{N}} \sim \mathcal{N}(\mathbf{0}, \mathbf{I})$ and then progressively denoising it under the guidance of the condition prompt $C$, following $\mathbf{S}_{t_{0}} = \mathcal{G}_S(\mathbf{S}_{t_{N}} |  C)$. The resulting ``clean'' latent voxel feature $\mathbf{S}_{t_{0}}$ is then decoded by a structure decoder $\mathcal{D}_S$ to obtain the occupied voxel coordinates $\mathbf{P}$, i.e., $\mathbf{P}=\mathcal{D}_S(\mathbf{S}_{t_{0}})$. 

In the second stage, conditioned on the generated voxel coordinates $\mathbf{P}$, the structured latent generator $\mathcal{G}_L$ synthesizes the corresponding latent features
$\mathbf{F}$ from an initial Gaussian noise 
$\mathbf{F}_{t_N} \sim \mathcal{N}(\mathbf{0}, \mathbf{I})$, following $\mathbf{F}_{t_{0}} = \mathcal{G}_L(\mathbf{F}_{t_{N}} | \mathbf{P} , C)$. The SLat representation $\mathbf{Z}$ is therefore constituted via $ \mathbf{Z} = \{ \mathbf{P}, \mathbf{F}\}$, which is next passed through an object decoder $\mathcal{D}$ to be converted into a 3D asset $A$ in diverse target representations, e.g., NeRF, 3D Gaussian Splatting, or mesh, via $A = \mathcal{D}(\mathbf{Z}_{t_0})$. $\mathcal{G}_{S}$ and $ \mathcal{G}_{L}$ are trained separately using the CFM objective.

\section{Methods}
Let us now introduce our method, \textbf{OrnaStyler}, to style editing for 3D assets. Formally, given a content 3D asset $A^c$ derived from the text description $D^c$, our goal is to generate an edited asset $A^e$ guided by a new text prompt $D^e$. Here, the edited prompt $D^e$ fully retains the original content description $D^c$ but entails an additional style description $D^s$ (i.e., $D^e = \{ D^c, D^s\}$). As such, $A^e$ is expected to be consistent with the style specified in $D^s$, while faithfully preserving the geometric structure and appearance cues of the original asset $A^c$. Unlike previous approaches that directly edit the explicit 3D representations or require full regeneration of geometry, 
we propose to explicitly model the spatial distribution of the ornaments with flow inversion, and then harmonize the appearance between style and content with 3D inpainting. As depicted in Fig. \ref{fig: pipeline}, our editing pipeline is organized in two stages: (i) \textbf{Voxel-level Ornament Synthesis} and (ii) \textbf{Latent-level Style Editing}, with each stage corresponding to the procedure in producing SLat.

\subsection{Flow Inversion-Guided Ornament Stylization} \label{sec: voxel-level}
While prior techniques (e.g., TRELLIS) can perform style editing by directly regenerating voxels using the edited text $D^e$, such a strategy often leads to significant content degradation, as the original shape $A^c$ is not explicitly preserved. To maintain strong content fidelity, we instead perform editing directly on the latent embeddings of the content voxels $\mathbf{P}^c$, and then spatially localize the active voxel set $\mathbf{P}^e$ corresponding to the edited asset $A^e$. Since the voxel coordinates 
$\mathbf{P}^c$ are generated via RF, we are naturally motivated to leverage flow inversion to recover their latent representations to facilitate editing.

\noindent \textbf{Flow Inversion.} In general, flow inversion aims to reconstruct the initial latent noise by reversing the ordinary differential equation (ODE) trajectory of deterministic samplers. Formally, let $\hat{x}_t$ be a latent in the inversion process. Following the step-wise denoising described in Eq. \ref{eq:flow_euler_solver}, for the flow model with an Euler method solver, the inverted latent $\hat{x}_{t_{i+1}}$ can be derived as 
\begin{equation}
\hat{x}_{t_{i+1}} = \hat{x}_{t_i} - (t_{i} - t_{i+1})\bm{v}(\hat{x}_{t_{i+1}},t_{i+1}, C),
\label{eq:inversion1}
\end{equation}
where $C$ is similarly included for conditioning. While Eq. \ref{eq:inversion1} formulates a principled inversion paradigm, as $\hat{x}_{t_{i+1}}$ is unknown in each step, $\bm{v}(\hat{x}_{t_{i+1}},t_{i+1}, C)$ remains inaccessible.
To nonetheless make the inversion tractable, we refer to \cite{jiao2026unieditflow} to approximate the unknown $\hat{x}_{t_{i+1}}$ by preparing a proxy latent $\tilde{x}_{t_{i+1}}$. Specifically, we first estimate $\tilde{x}_{t_{i+1}}$ using the velocity from the previous step $\hat{\bm{v}}_{t_i}$:
\begin{equation}
\tilde{x}_{t_{i+1}} = \hat{x}_{t_i} - (t_{i} - t_{i+1})\hat{\bm{v}}_{t_i},
\label{eq:inversion2}
\end{equation}
and then replace the inaccessible $\bm{v}(\hat{x}_{t_{i+1}},t_{i+1}, C)$ in Eq. \ref{eq:inversion1} with $\hat{\bm{v}}_{t_{i+1}} = \bm{v}(\tilde{x}_{t_{i+1}}, t_{i+1}, C)$ to perform the actual update:
\begin{equation}
\hat{x}_{t_{i+1}} = \hat{x}_{t_i} - (t_{i} - t_{i+1})\hat{\bm{v}}_{t_{i+1}}.
\label{eq:inversion3}
\end{equation}
The stable backward tracing of the latent trajectory can therefore be achieved by alternately calling Eq. \ref{eq:inversion2} and Eq. \ref{eq:inversion3}.

Since we aim to derive the inverted latent of the content structure, we endow $\hat{x}_0$ with the latent voxel embedding $\mathbf{S}^c$ derived with the structure encoder $\mathcal{E}_S$ (i.e., $\mathbf{S}^c=\mathcal{E}_S(\mathbf{P}^c)$), and $C$ with the source text $D^c$, respectively, to initialize the inversion. A series of inverted latents $\{\hat{\mathbf{S}}^{c}_{t_{i}}\}_{i=\{1,\cdots,N\}}$ can therefore be obtained. In practice, we notice that fully inverting the latent to pure noise $\hat{\mathbf{S}}^{c}_{t_{N}}$ tends to erase fine-grained geometric details of the original shape. To combat this tendency, we introduce a control parameter $\alpha \in (0,1]$ to 
regulate the inversion depth (i.e., delayed injection \cite{xiao2025fastcomposer}), and extract an intermediate latent $\hat{\mathbf{S}}^{c}_{t_{\alpha N}}$. In principle, this partially inverted latent preserves the core structure of $\mathbf{P}^c$. We next need to determine how to generate stylized voxels $\mathbf{P}^e$ with desired ornament shapes by using $\hat{\mathbf{S}}^{c}_{t_{\alpha N}}$ as the seed shape latent.


\noindent \textbf{Region-Aware Ornament Stylization.} A naive way to produce $\mathbf{P}^e$ is to directly apply the structure generator $\mathcal{G}_S$ using $\hat{\mathbf{S}}^{c}_{t_{\alpha N}}$ and the edited text prompt $D^e$. However, we observe that this may introduce conflicts between the original content structure and the newly imposed style. This is because stylistic geometry often appears in the form of ornaments, whose spatial layout is intrinsically constrained by the underlying content shape. For example, ornament elements, like vines, should be located in accordance with the underlying surface geometry.
Also, conditioning solely on the styled text prompt $D^e$ provides ambiguous guidance in content regions, which suppresses the generation of delicate and localized ornament structures.
While applying geometric constraints (e.g., SDF-based physical regularizations) can provide more direct guidance, they can also induce extra computational overhead.

To address the above issue, we propose a lightweight region-aware structure stylization scheme that adaptively controls the synthesis of ornamented voxels during denoising. 

Specifically, for any content voxel embedding $\mathbf{S}^c$, we define an occupancy mask $M^o = \{m^o_u\}_{u=1}^{U}$ over the structure-embedding voxel grid of capacity $U$ to distinguish the occupied content regions from the unoccupied regions for ornament synthesis, whose spatial entries are given by
\begin{equation}
\label{eq: structured mask}
m^o_u = 
\begin{cases}
    1 & \text{if $u \in L'$} \\
    0 & \text{otherwise},
\end{cases}
\end{equation}
where $L'$ denotes the active content voxels in the embedding space. Intuitively, ornamented voxels should only be activated in regions where $m^o_u=0$.

Given $M^o$, we individually compute two velocity fields using $\mathcal{G}_S$, by conditioning one on the original content prompt and the other on the edited prompt.
These velocities are then fused in a region-adaptive manner as
\begin{equation}
\tilde{\bm{v}}^{{s}}_{t_i} = M^o \odot \bm{v}^s(\mathbf{S}^s_{t_i}, t_i, D^c) + (1 - M^o) \odot \bm{v}^s(\mathbf{S}^s_{t_i}, t_i, D^e),
\label{eq:tiled_v^s}
\end{equation}
where $\bm{v}^s$ denotes the denoising velocity predicted by $\mathcal{G}_S$, $\odot$ represents the element-wise product, and $\mathbf{S}^s_{t_i}$ is the structure embedding at timestep $t_i$. Concretely, the $D^c$-conditioned velocity plays the role of a corrective force within the content regions to prevent excessive structural distortion, while the $D^e$-conditioned velocity enables stylistic augmentation in non-content regions.
We additionally introduce a style-enhancement correction based on the weighted
difference between the edited- and source-conditioned velocities, following
\cite{jiao2026unieditflow}:
$\bm{v}^{\mathrm{cfg}}_{t_i}
= w_s\big[\bm{v}^s(\mathbf{S}^s_{t_i},t_i,D^e)-\bm{v}^s(\mathbf{S}^s_{t_i},t_i,D^c)\big]$.
Here, we incorporate it as a masked correction to obtain
$\bar{\mathbf{S}}^s_{t_i}=\mathbf{S}^s_{t_i}+(t_{i-1}-t_i)M^o \odot\bm{v}^{\mathrm{cfg}}_{t_i}$, which aims to spatially anchor the content regions to the source.
The update of the structure embedding is eventually formulated as
\begin{equation}
\mathbf{S}^s_{t_{i-1}} =  \bar{\mathbf{{S}}}^s_{t_{i}} + (t_{i-1} - t_{i})\tilde{\bm{v}}^{{s}}_{t_i},
\label{eq: S^s_t_i-1}
\end{equation}
with initialization $\mathbf{S}^{s}_{t_{\alpha N}} = \hat{\mathbf{S}}^{c}_{t_{\alpha N}} $.

Then, the denoised output $\mathbf{S}^s_{t_0}$ is again projected back to the voxel space using the decoder $\mathcal{D}_S$. To resolve potential structural discontinuities or holes, we explicitly merge the decoded voxels with the original content voxels to create the stylized structure $\mathbf{P}^e$, following $\mathbf{P}^e =  \mathcal{D}_S(\mathbf{S}^s_{t_0}) \cup \mathbf{P}^c $. $\mathbf{P}^e =\{ p_{h}\}^{H}_{h = 1}$ eventually contains $H$ active voxels. \textit{In essence, $\mathbf{P}^e$ stores the geometric occupancy for ornamented content shape described in the style cues}. We next need to determine how to create the corresponding latent feature $\mathbf{F}^e$ to compose the SLat.

\subsection{Latent-Level Style Editing} \label{sec: latent-level style editing}
Directly feeding $\mathcal{G}_L$ with the edited voxel coordinates 
$\mathbf{P}^e$ and prompt $D^e$ can generate the target latent feature 
$\mathbf{F}^e$. However, this can damage the original content appearance encoded in $\mathbf{F}^c$. To preserve content identity, we design a feature-level editing strategy that mirrors the voxel-level editing paradigm by explicitly maintaining content characteristics. Specifically, we first stylize the features associated with the content voxels to obtain style-aware content features, and subsequently harmonize them with ornament features in the manner of 3D inpainting.

\noindent \textbf{Content Appearance Editing.}
As in the voxel-level editing, we first identify the feature seed latent of $\mathbf{F}^c$. In particular, given the encoded content feature $\mathbf{F}^c$ and $\mathbf{P}^c$, its corresponding text prompt $D^c$, and style text prompt $D^s$, we again enforce flow inversion with the latent generator $\mathcal{G}_L$ to recover the initial noise latent $\hat{\mathbf{F}}^c_{t_N}$. 
Note that, in contrast to the voxel-level inversion, we invert the full trajectory without any delaying, since the geometric structure has already been fixed in $\mathbf{P}^c$ and encoded within $\mathbf{F}^c$ to avoid structural distortion. 
We then create the style-enhanced feature $\mathbf{F}^{cs}$, following the CFM with $\mathcal{G}_L$:
\begin{equation}
\mathbf{F}^{cs} =\mathcal{G}_{L}(\hat{\mathbf{F}}^c_{t_N} \mid \mathbf{P}^c, D^{e}),
\label{eq:s^o}
\end{equation}
which injects stylistic cues while maintaining consistency with the original content appearance to facilitate editing with $\mathbf{F}^{cs}$.

\begin{figure}[t]
    \centering
    \includegraphics[width=\linewidth]{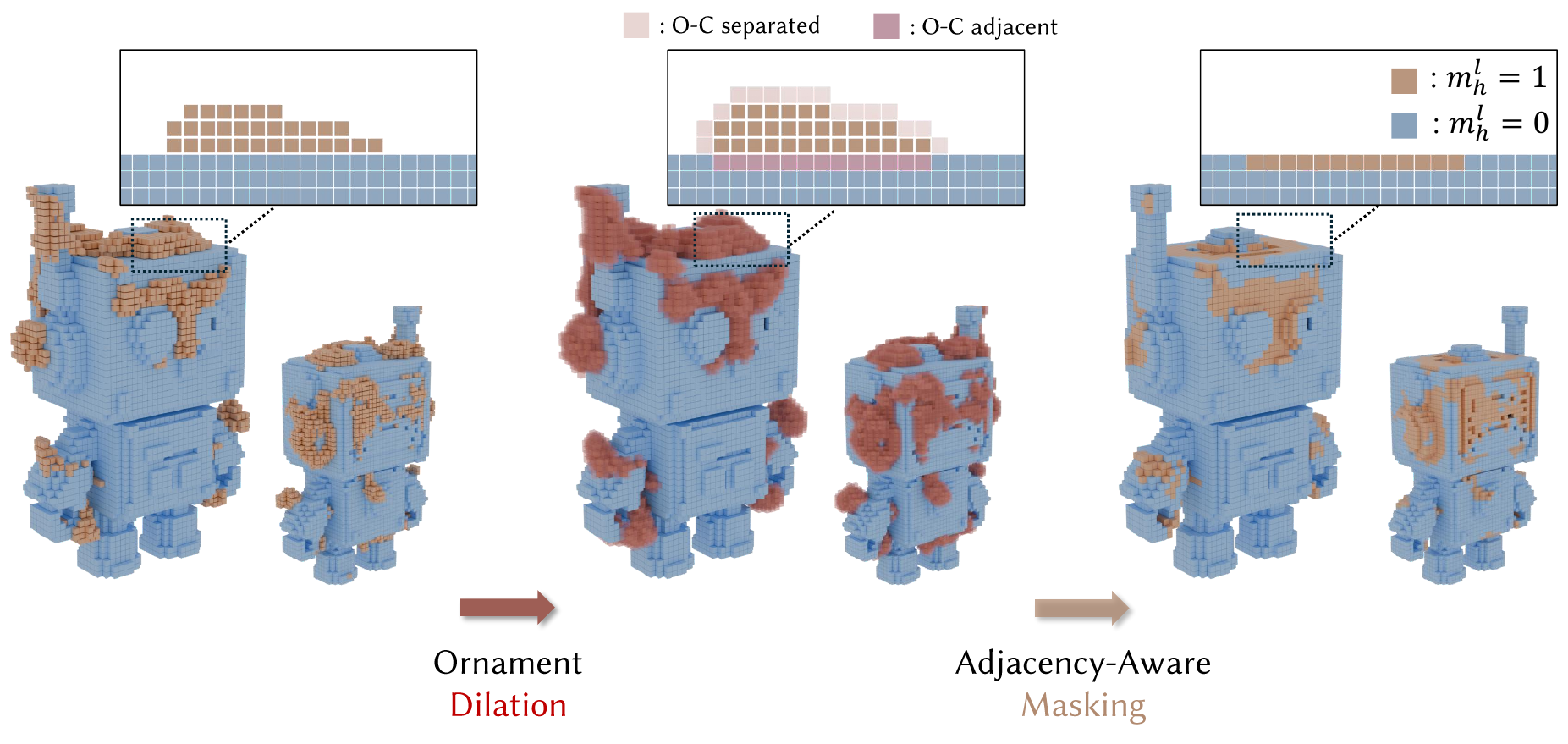}
    \caption{\textbf{An example of adjacency-aware masking}, which is progressively constructed by determining the overlapping area between the content and dilated ornament voxels. O-C refers to Ornament-Content.  }
    \label{fig: adj_mask}
\end{figure}

\noindent \textbf{Adjacency-Aware Masking.}
Although $\mathbf{F}^{cs}$ provides stylized representations for the original content voxels, the appearance of newly generated ornament voxels needs to be inferred in a way that remains coherent with surrounding structures. We are thus motivated to explicitly analyze the spatial adjacency between content and ornament regions to navigate feature synthesis and integration. To this end, we incorporate a feature-level inpainting mechanism $\mathcal{G}_L$, which is further guided by a spatial mask $M^l=\{m^l_{h} \}^H_{h=1}$ that captures adjacency between content and ornament voxels, to generate the final latent feature $\mathbf{F}^{e}$.

Specifically, to define the target region, we first isolate the ornament coordinates $\mathbf{P}^s$ by subtracting $\mathbf{P}^c$ from $\mathbf{P}^e$: $
\mathbf{P}^s = \mathbf{P}^e \setminus \mathbf{P}^c
$. We then perform 3D morphological dilation on $\mathbf{P}^s$ to determine the regions of influence around these ornament voxels, following $\mathbf{P}^{d} = \text{Dil}_{\eta}(\mathbf{P}^s)$, where $\text{Dil}_{\eta}(\cdot)$ denotes the dilation operator with $\eta$ parameterizing the iteration number in operating. Eventually, to identify the adjacent region, we measure the intersection of $\mathbf{P}^e$ and $\mathbf{P}^c$, and then union it with the original ornament coordinates $\mathbf{P}^s$. The entries of our spatial mask $m^l_h$ can thus be given by 
\begin{equation}
\label{eq:eq6}
m^l_h = 
\begin{cases}
    1 & \text{if $h \in  (\mathbf{P}^d \cap \mathbf{P}^c)  \cup \mathbf{P}^s$} \\
    0 & \text{otherwise},
\end{cases}
\end{equation}
where $(\mathbf{P}^d \cap \mathbf{P}^c)$ denotes the active adjacent region in content voxels. An example of the construction of our spatial mask is displayed in Fig. \ref{fig: adj_mask}. Precisely, $M^l$ delineates where feature integration should occur to enable seamless synthesis between content and ornament regions for stylization. 
Unlike the occupancy mask $M^o$ defined in the structure-embedding space, $M^l$ is constructed after structure decoding over the edited voxel coordinates $\mathbf{P}^e$ to identify the ornament voxels and their adjacent content regions for feature integration.

\begin{figure*}
    \centering
    \includegraphics[width=\linewidth]{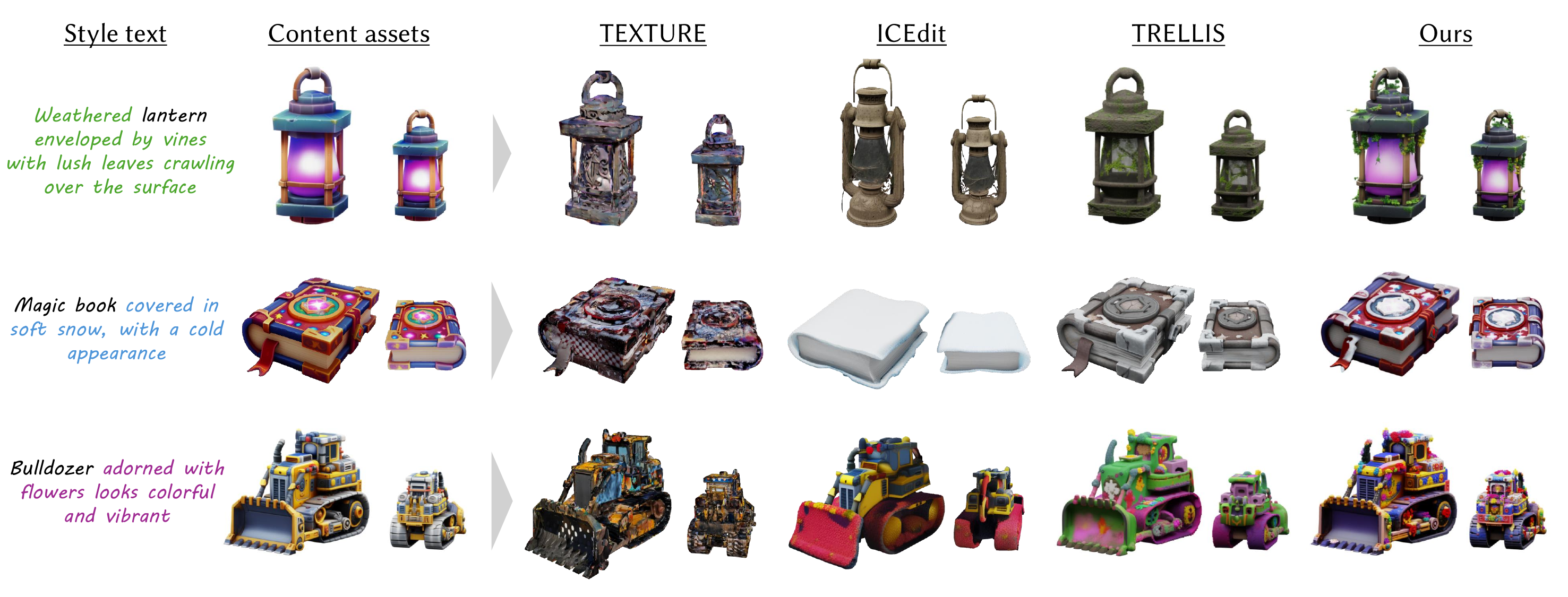}
    \caption{\textbf{Qualitative results} against prior style editing approaches. \textcolor{OliveGreen}{Colored}  \textcolor{NavyBlue}{textual}  \textcolor{RedViolet}{prompts} denote the style texts for editing the content asset generated with the black texts. }
    \label{fig: quantitative_comparision}
\end{figure*}

\noindent \textbf{Inpainting-Inspired Feature Harmonization.}
To generate the final latent $\mathbf{F}^{e}$, we are inspired by \cite{lugmayr2022repaint} to extend a zero-shot inpainting strategy for 3D inpainting to achieve harmonized stylization. 
Our goal here is to synthesize local latent features for the edited coordinates $\mathbf{P}^e$ within the spatial mask $M^l$, while preserving the integrity of the edited content feature $\mathbf{F}^{cs}$.

Specifically, given the edited voxel coordinates $\mathbf{P}^e$, we first perform feature generation via CFM using the latent generator $\mathcal{G}_L$ by initializing $\mathbf{F}^e_{t_N}$ with random noise. Following the reverse flow dynamics, the latent features are iteratively updated by estimating the velocity $\bm{v}^f_{t_i} = \bm{v}^f(\mathbf{F}^e_{t_i}, t_i, \mathbf{P}^e, D^s)$, conditioned on the ornament design text $D^s$:
\begin{equation}
\tilde{\mathbf{F}}^e_{t_{i-1}} = \mathbf{F}^e_{t_{i}} + (t_{i-1} - t_{i})\bm{v}^f_{t_{i}}.
\label{eq:update_latent_harmoni}
\end{equation}
To preserve consistency with the stylized content features, we then re-noise the features for the content region using linear interpolation between the known data $\mathbf{F}^{cs}$ and the target distribution ($\mathcal{N}(\mathbf{0}, \mathbf{I})$):
\begin{equation}
\psi_{t_{i-1}}(\mathbf{F}^{cs})=(1-t_{i-1})\mathbf{F}^{cs}+[\sigma_{min}+(1-\sigma_{min})t_{i-1}]\bm{\epsilon},
\label{eq:forward}
\end{equation}
where $\sigma_{min}$  denotes the minimum noise scale used by the pretrained rectified model and $\bm{\epsilon} \sim \mathcal{N}(\mathbf{0}, \mathbf{I})$.
By incorporating the masking cues within $M^l$, the update for $\mathbf{F}^e_{t_i}$ is formalized with
\begin{equation}
\mathbf{F}^e_{t_{i-1}} = M^l \odot \tilde{\mathbf{F}}^e_{t_{i-1}} + (1 - M^l) \odot \psi_{t_{i-1}}(\mathbf{F}^{cs}).
\label{eq:inpaint_feature}
\end{equation}
Since $M^l$ captures dense adjacency, the iterative denoising with Eq. \ref{eq:inpaint_feature} gradually inpaints the style (i.e., ornament) appearance in a content-harmonized manner. The denoised $\mathbf{F}^e_{t_0}$ is therefore used to constitute the target SLat $\mathbf{Z}^e = \{\mathbf{P}^e, \mathbf{F}^e_{t_0}\}$, which is then fed to the object decoder $\mathcal{D}$ to generate the final stylized 3D asset $A^e$. \textit{See Appendix~\ref{appn: algorithm} for a detailed algorithm for both stages of our method.}

\section{Experiments}
In this section, we conduct extensive experiments to evaluate the effectiveness of our method against existing text-based 3D style editing methods.

\noindent \textbf{Implementation Details.}
As OrnaStyler is by design zero-shot, we directly incorporate the state-of-the-art text-driven 3D generation model from TRELLIS \cite{xiang2025structured} (i.e., pre-trained TRELLIS-text-xlarge) as the backbone. Following TRELLIS, ($d$, $K$) is set to (8, 64). During editing, we uniformly set the number of denoising integration steps to 25 for both generators $\mathcal{G}_S$ and $\mathcal{G}_L$. 
The parameters for the delayed rate and dilation number ($\alpha$, $\eta$) are set to (0.8, 1).

\begin{table*}
    \centering
    \renewcommand{\arraystretch}{1.0}
    \caption{\textbf{Quantitative evaluation} of editing performance in terms of content-consistency on the TRELLIS-based and Sketchfab datasets. }
    \resizebox{\linewidth}{!}{
        \begin{tabular}{lcccccccccc}
            \toprule
                                                     \textbf{Method} & \multicolumn{5}{c}{\textbf{TRELLIS-Generated} }
                                                     & \multicolumn{5}{c}{\textbf{Sketchfab}} \\
                                                     \cmidrule(lr){2-6} \cmidrule(lr){7-11}
                                                    &{SSIM ↑} &{PSNR ↑}&{CLIP ↑}&{LPIPS ↓}&  {FID ↓} &{SSIM ↑} &{PSNR ↑}&{CLIP ↑}&{LPIPS ↓}&  {FID ↓}\\  \hline 
            Text2Tex \textit{\fontsize{7}{10}\selectfont[ICCV'23]}        &   0.761 &  16.42 &  0.841 &  0.205  &  120.4 
                                                    &   0.844 &  18.38 &  0.872 &  0.168  &  124.6 \\ 
            TEXTure \textit{\fontsize{7}{10}\selectfont[SIGGRAPH'23]}     &   0.726 &  16.04 &  0.815 &  0.217  &  151.1  
                                                    &   0.805 &  18.08 &  0.816 &  0.174  &  174.6 \\ 
            TexPainter \textit{\fontsize{7}{10}\selectfont[SIGGRAPH'24]}    &   0.769 &  17.12 &  0.837 &  0.210  &  132.9  
                                                    &   0.707 &  13.91 &  0.830 &  0.268  &  143.3 \\ 
            SyncMVD \textit{\fontsize{7}{10}\selectfont[SIGGRAPH'24]}               &   0.752 &  16.01 &  0.838 &  0.206  &  118.2  
                                                    &   0.822 &  17.32 &  0.860 &  0.173  &  136.4 \\ 
            Paint3D \textit{\fontsize{7}{10}\selectfont[CVPR'24]}           &   0.752 &  15.73 &  0.857 &  0.203  &  117.5  
                                                    &   0.839 &  17.01 &  0.884 &  0.164  &  115.4 \\   
            TRELLIS \textit{\fontsize{7}{10}\selectfont[CVPR'25]} + ICEdit \textit{\fontsize{7}{10}\selectfont[NeurIPS'25]}           &   0.643 &  13.02 &  0.793 &  0.289  &  125.5  
                                                    &   0.640 &  12.45 &  0.733 &  0.291  &  186.4 \\ 
            TRELLIS \textit{\fontsize{7}{10}\selectfont[CVPR'25]} + Gemini        &   0.653 &  13.46 &  0.788 &  0.299  &  167.0   
                                                    &   0.673 &  13.12 &  0.758 &  0.292  &  221.9\\ 
            TRELLIS \textit{\fontsize{7}{10}\selectfont[CVPR'25]}      &   0.766 &  17.54 &  0.863 &  0.197  &  106.4  
                                                    &   0.848 &  19.42 &  0.876 &  0.158  &  123.7 \\ 
            \rowcolor{OursBlue}
            \textbf{OrnaStyler}                                   &   \textbf{0.793} &  \textbf{19.36} &  \textbf{0.899} & \textbf{0.146}   &  \textbf{79.6} 
                                                    &   \textbf{0.885} &  \textbf{21.95} &  \textbf{0.898} &  \textbf{0.105}  &  \textbf{103.0} \\               
            
            \bottomrule
        \end{tabular}
    }
    \label{tb:score_sim}
\end{table*}

\noindent \textbf{Dataset.}
To evaluate the quality and quantity of the style-edited 3D assets, we collect 30 assets with diverse geometric and texture styles using TRELLIS. In particular, we condition the generation with either public images with explicit 3D structure or apply text descriptions annotated in the Objaverse-XL dataset \cite{deitke2023objaverse}. Furthermore, to assess the robustness of our method, we additionally collect 20 assets from Sketchfab\footnote{https://sketchfab.com. All asset attributions are listed in Appendix~\ref{appn: sketchfab}.} under Creative Commons Licenses. As our editing targets ornament-aware stylization, for editing texts, we prepare five distinct styles: vines, snow, seabed, flower, and mud, for each 3D content via text prompts.

\subsection{Evaluation of style adaptation}

\noindent \textbf{Qualitative Evaluation.}
We first evaluate the style editing performance of our method via visual comparison with existing methods. For a comprehensive assessment, we compare our method against two representative categories of state-of-the-art text-guided 3D editing paradigms: \textit{2D-edit-3D-generation methods and \textit{UV-texture-based methods}.} For the \textit{2D-edit-3D-generation} paradigm, we first edit rendered images of the content assets using a state-of-the-art image editing model (i.e., ICEdit \cite{zhang2025enabling}), and subsequently reconstruct the edited 3D assets using TRELLIS \cite{xiang2025structured}. For the \textit{UV-texture-based} paradigm, we compare against TEXTure \cite{richardson2023texture} and Paint3D \cite{zeng2024paint3d}. Also, we compare with TRELLIS \cite{xiang2025structured} under its editing mode. Fig. \ref{fig: quantitative_comparision} visualizes the comparison. 
It can be observed that while UV-texture editing methods \cite{richardson2023texture, zeng2024paint3d} inherently preserve the content geometry (i.e., the mesh), they often struggle to maintain design consistency with the original content. Moreover, the resulting textures tend to exhibit limited responsiveness to different style prompts, leading to visually similar outputs despite varying textual descriptions. For \textit{2D-edit-3D-generation} approaches, while they generally produce stylistically plausible images, the reconstructed 3D assets can suffer from structural deviations due to the domain gap between 2D edits and 3D geometry, particularly in occluded regions such as the back side of objects.
In contrast, our OrnaStyler produces stylized 3D assets with higher visual fidelity, while maintaining stronger consistency with the original content geometry. Furthermore, the generated ornament structures are seamlessly integrated with the existing shapes, resulting in coherent and natural stylization.
\textit{More results are provided in Appendix.}

\noindent \textbf{Quantitative Evaluation.}
We next report the quantitative evaluation results against prior text-based 3D stylization methods. To assess the similarity between content assets and their edited counterparts, we render both assets from 12 fixed viewpoints and compute similarity scores for each view using four widely adopted metrics: Structure similarity (SSIM) \cite{wang2004image}, Peak signal-to-noise ratio (PSNR), Learned Perceptual Image Patch Similarity (LPIPS) \cite{zhang2018perceptual}, Contrastive Language-Image Pre-Training (CLIP) \cite{radford2021learning}. In addition, we compute Fréchet Inception Distance (FID) \cite{heusel2017gans} to measure distribution-level similarity between the rendered images of edited assets and those of the reference content assets. We individually evaluate on our collected datasets, the TRELLIS-generated and the Sketchfab dataset. For each content asset, we endow all five different editing prompts and then render the resulting edited asset, yielding 1,800 and 1,200 images for each dataset, respectively. The scores for each metric are averaged across all rendered views. We compare OrnaStyler with several representative baselines\footnote{We use the officially released pre-trained models for all baselines.}, Text2Tex  \cite{chen2023text2tex}, TEXTure \cite{richardson2023texture}, TexPainter \cite{zhang2024texpainter}, SyncMVD \cite{liu2024text}, Paint3D \cite{zeng2024paint3d}, TRELLIS \cite{xiang2025structured}, Gemini, and ICEdit \cite{zhang2025enabling}. For Gemini and ICEdit, we follow the same protocol as in the qualitative evaluation, where edited images are first generated and then reconstructed into 3D assets using TRELLIS. Additional details on rendering and evaluation protocols are provided in Appendix~\ref{sec: appendix_qualitative}. \textit{Note that the quantitative evaluation here primarily focuses on content preservation rather than style fidelity.}

The results are summarized in Tab. \ref{tb:score_sim}.  In general, UV-texture editing approaches \cite{richardson2023texture, zhang2024texpainter, zeng2024paint3d, liu2024text} achieve better content preservation than 2D image-based techniques \cite{zhang2025enabling, comanici2025gemini}, which also reflects in the visual comparisons in Fig. \ref{fig: quantitative_comparision}. Nevertheless, it can be confirmed that OrnaStyler consistently outperforms all the compared methods across all evaluation metrics. While the additional geometric structures introduced by ornamentation can potentially decrease structural similarity, our method still yields minimum content design leakage. We can thus verify that OrnaStyler achieves a better balance between content preservation and stylization compared to existing approaches.

\noindent \textbf{User Study.}
As there is no universally accepted metric for evaluating stylization quality, and human perception remains the most reliable criterion for assessing the editing quality, we conduct a user study to perceptually assess the editing quality. Following \cite{xiang2025structured}, we adopt a Single Best Answer protocol to compare our method against four representative baselines: \textit{UV-texture-based} editing approaches (i.e., TEXTure \cite{richardson2023texture}, Paint3D \cite{zeng2024paint3d}), \textit{2D-Edit to 3D-Generation} approach (i.e., Gemini \cite{comanici2025gemini}), and the base model (i.e., TRELLIS \cite{xiang2025structured}). For each content asset, we generate stylized results using identical text prompts across all methods. We randomly select 20 samples from the evaluation dataset and recruit 30 participants for the study. For each sample, participants are asked to choose the best result among different methods according to the following criteria:
(i) \textit{Style fidelity and content preservation}, which evaluates how well the style is applied while maintaining consistency with the original content appearance;
(ii) \textit{Quality of style ornamentation}, which measures the effectiveness and richness of style-specific ornament integration;
(iii) \textit{Overall naturalness}, which assesses the visual realism and plausibility of both geometry and appearance. The complete questionnaire and details of our user study are provided in Appendix~\ref{appn: user_study}.

The statistics of the perceptual study are tabulated in Tab. \ref{tb:user_study}. Overall, our method is consistently preferred perceptually by participants over competing approaches across all evaluation criteria. In particular, while the LLM-based method (i.e., Gemini) demonstrates strong capability in producing visually appealing 2D edits, its powerful generative nature often leads to significant deviations from the original content geometry after 3D reconstruction. By contrast, as our method is specifically designed to prioritize ornaments, it harvests higher preference scores in both style fidelity and overall naturalness, indicating the effectiveness in producing more visually plausible and structurally faithful 3D edits.

\begin{table}[t]
    \centering
    \renewcommand{\arraystretch}{1.0}
    \caption{\textbf{Statistic from user study.} Each value represents the percentage (\%) of participant preferences among five methods for (i) style fidelity and content preservation, (ii) quality of style ornamentation, and (iii) overall naturalness.}
    \resizebox{\linewidth}{!}{
        \begin{tabular}{lccc}
            \toprule
                                                    \textbf{Method} & \textbf{(i): Preservation } &  \textbf{(ii): Ornamentation } &  \textbf{(iii): Naturalness}  \\  \hline 
            TEXTure \textit{\fontsize{7}{10}\selectfont[SIGGRAPH'23]}     &  9.3  &  10.5  &   12.9 \\ 
            Paint3D \textit{\fontsize{7}{10}\selectfont[CVPR'24]}         &  5.7  &  4.7   &   6.3 \\ 
            Gemini        &  21.8  &  30.1   &   24.0\\ 
            TRELLIS \textit{\fontsize{7}{10}\selectfont[CVPR'25]}       &  15.0  &  11.7   &   16.0 \\ 
            \rowcolor{OursBlue}
            \textbf{OrnaStyler}                                     &  \textbf{48.2}   &  \textbf{43.0}  & \textbf{40.8} \\ 
            \bottomrule
        \end{tabular}
    }
    
    \label{tb:user_study}
\end{table}

\begin{figure}[]
    \centering
    \includegraphics[width=\linewidth]{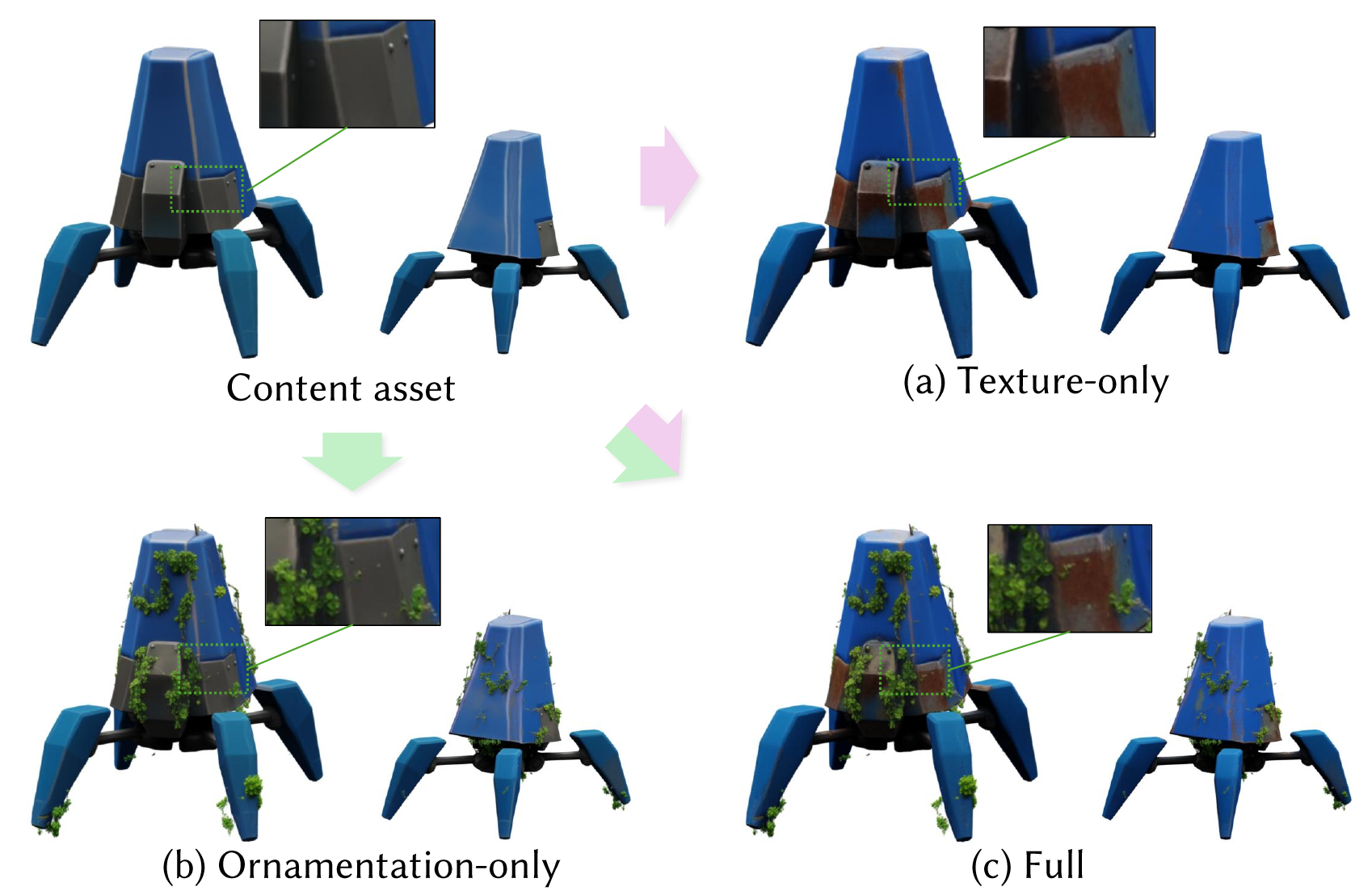}
    \caption{\textbf{Selective editing results.} For the content asset, our method allows for (a) texture-only editing, (b) ornamentation-only editing, and (c) full editing.}
    \label{fig: qualitative_analysis}
\end{figure}

\begin{table}[t]
    \centering
    \renewcommand{\arraystretch}{1.0}
    \caption{\textbf{Quantitative evaluation of content consistency under different editing modes} on Sketchfab. CD is multiplied by $10^3$.}
    \resizebox{0.8\linewidth}{!}{
        \begin{tabular}{lcccc}
            \toprule
                                                     \textbf{Setting} & \multicolumn{2}{c}{\textbf{Voxel-level}} 
                                                     & \multicolumn{2}{c}{\textbf{Asset-level}} \\
                                                      \cmidrule(lr){2-3} \cmidrule(lr){4-5}
                                                     &{IoU ↑}&  {CD ↓} &{LPIPS ↓}&  {FID ↓}\\  \hline 
             
            Texture-only                       & 1.0   &  0.0  
                                                    & 0.083  &  81.3 \\ 
            Ornament-only                      & 0.897  &  0.103
                                                    & 0.075  &  44.8 \\ 
            Full                               & 0.897  &  0.103
                                                    & 0.105  &  103.0 \\

            \bottomrule
        \end{tabular}
    }
    \label{tb:score_sim_analysis}
\end{table}

\subsection{Evaluation of controlling capacity}
OrnaStyler further enables flexible editing controls, spatially and semantically. Below we study each case.

\noindent \textbf{Selective Editing.}
As our method follows a two-stage design, we can decouple the editing process and enforce each stage independently. This allows us to apply two different editing modes: (i) \textit{Texture-only stylization}, which is achieved by disabling the voxel-level ornament generation stage; (ii) \textit{Ornamentation-only stylization}, which is achieved by substituting the content appearance feature $\mathbf{F}^c$ for the edited appearance feature  $\mathbf{F}^{cs}$ during the latent-level inpainting (Eq. \ref{eq:inpaint_feature}) to only add ornament geometry. The results are visualized in Fig.~\ref{fig: qualitative_analysis}. We observe that each mode contributes to distinct aspects of stylization. For mode (i) (Fig.~\ref{fig: qualitative_analysis}(a)), our method focuses solely on appearance adaptation (e.g., weathering effects) without introducing geometric distortion, while for mode (ii) (Fig.~\ref{fig: qualitative_analysis}(b)), the introduced ornament structures are spatially coherent with the underlying geometry with the original appearance unchanged. Combining both modes yields the full editing capability (Fig.~\ref{fig: qualitative_analysis}(c)). We can thus confirm the ability of OrnaStyler to effectively disentangle geometric augmentation from appearance stylization, enabling flexible and selective editing control.

\noindent \textbf{Influence of Editing Modes.}
To further examine the editing mode discussed in the last paragraph, we quantitatively analyze the influence of each mode on content preservation. Here, we compare the results of each mode (i.e., (i) Texture-only,  (ii) Ornamentation-only, and full) with the content shape on both voxel- and asset-level. The voxel-level comparison adopts the 3D Intersection-over-Union (IoU) and Chamfer Distance (CD), while for the asset level, we use LPIPS and FID to evaluate on the 2D image domain. The results on the Sketchfab dataset are presented in Tab. \ref{tb:score_sim_analysis}.
We observe that the texture-only mode strictly preserves the content geometry to achieve the best performance at the voxel level. Introducing ornamentation (i.e., modes (ii) and full) moderately reduces voxel-level geometric similarity, as new structures are included. Yet, the ornament-only mode still maintains strong visual coherence at the asset level, indicating that the introduced geometries remain perceptually consistent with the original content. Overall, there exists a trade-off between geometric fidelity and visual expressiveness. Nonetheless, OrnaStyler consistently yields superior content preservation compared to prior methods (as also shown in Tab.~\ref{tb:score_sim}), demonstrating its effectiveness in balancing geometric integrity and stylization.

\begin{figure}[t]
    \centering
    \includegraphics[width=\linewidth]{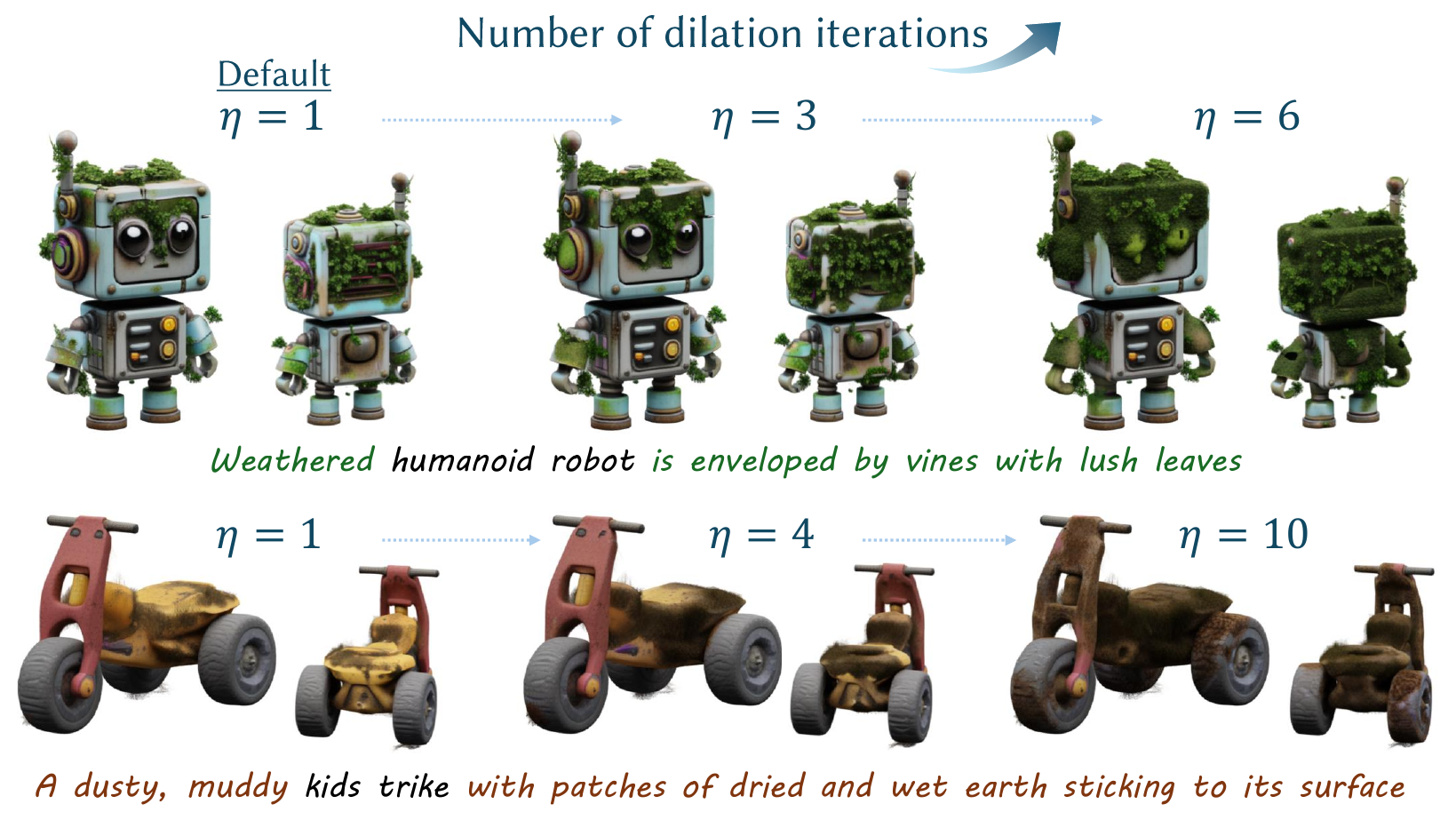}
    \caption{\textbf{Style intensity controlling} by varying dilation numbers. }
    \label{fig: mask_dilation}
\end{figure}

\noindent \textbf{Style Intensity Control.} 
As we employ morphological dilation to construct the adjacency-aware mask between content and ornament regions, it naturally provides a straightforward scheme for controlling style intensity. By varying the number of dilation iterations (i.e., $\eta$), we can adjust the spatial coverage over which ornament-related features are propagated and harmonized with the source asset. As shown in Fig.~\ref{fig: mask_dilation}, a larger dilation number expands the affected regions and yields stronger stylization effects. This demonstrates that OrnaStyler allows users to conveniently manipulate the strength of ornamental elements to yield intensity-controllable stylization. Also, it is worth noting that depending on the initial geometry and the textual description, the same increment in dilation iterations may result in non-uniform degrees of spatial expansion. 

\begin{figure}[t]
    \centering
    \includegraphics[width=0.9\linewidth]{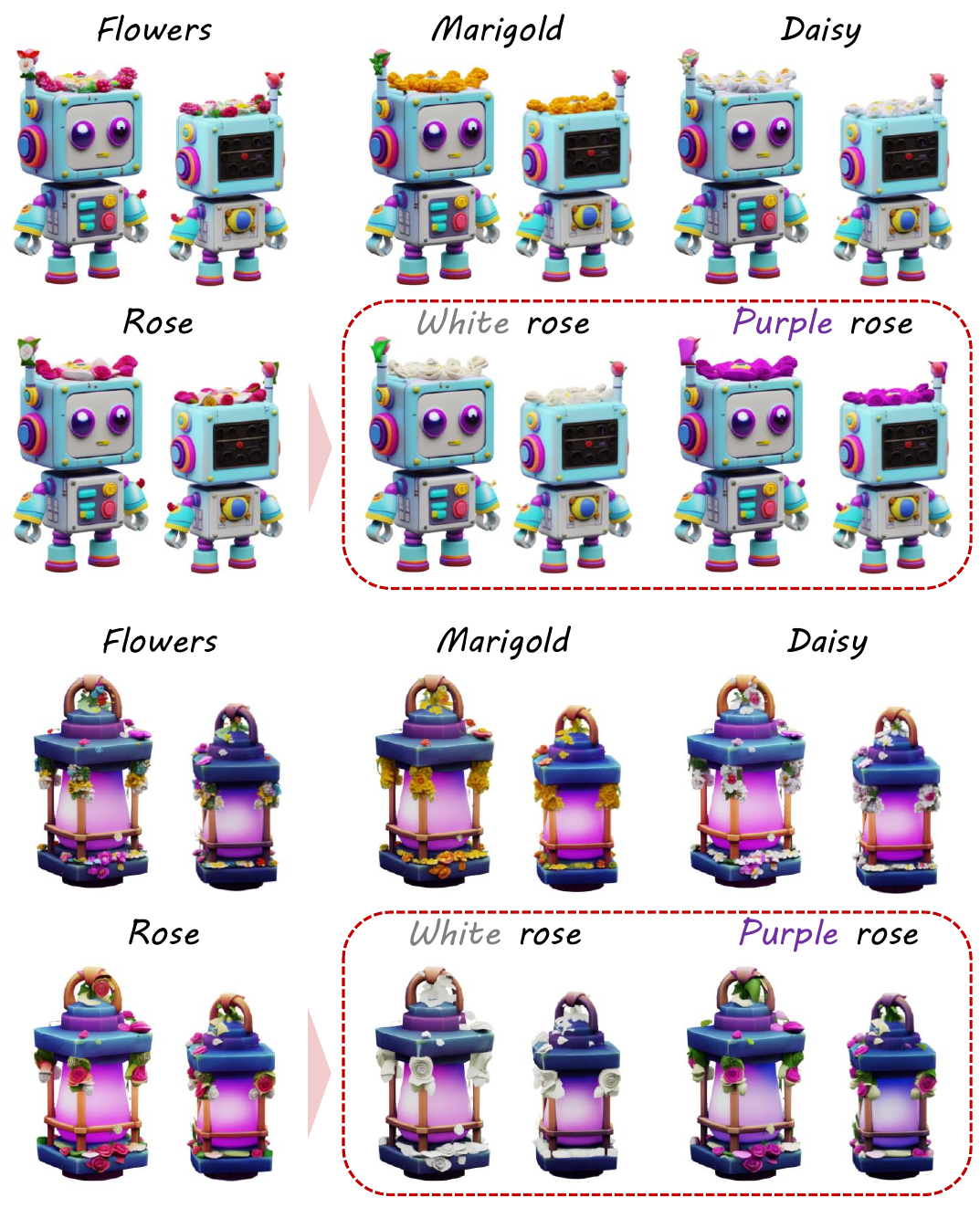}
    \caption{\textbf{Semantic controlling} over ornamentation regarding specific identity (e.g., rose types) or attributes (e.g., colors). }
    \label{fig: flowers}
\end{figure}

\begin{figure*}[]
    \centering
    \includegraphics[width=\linewidth]{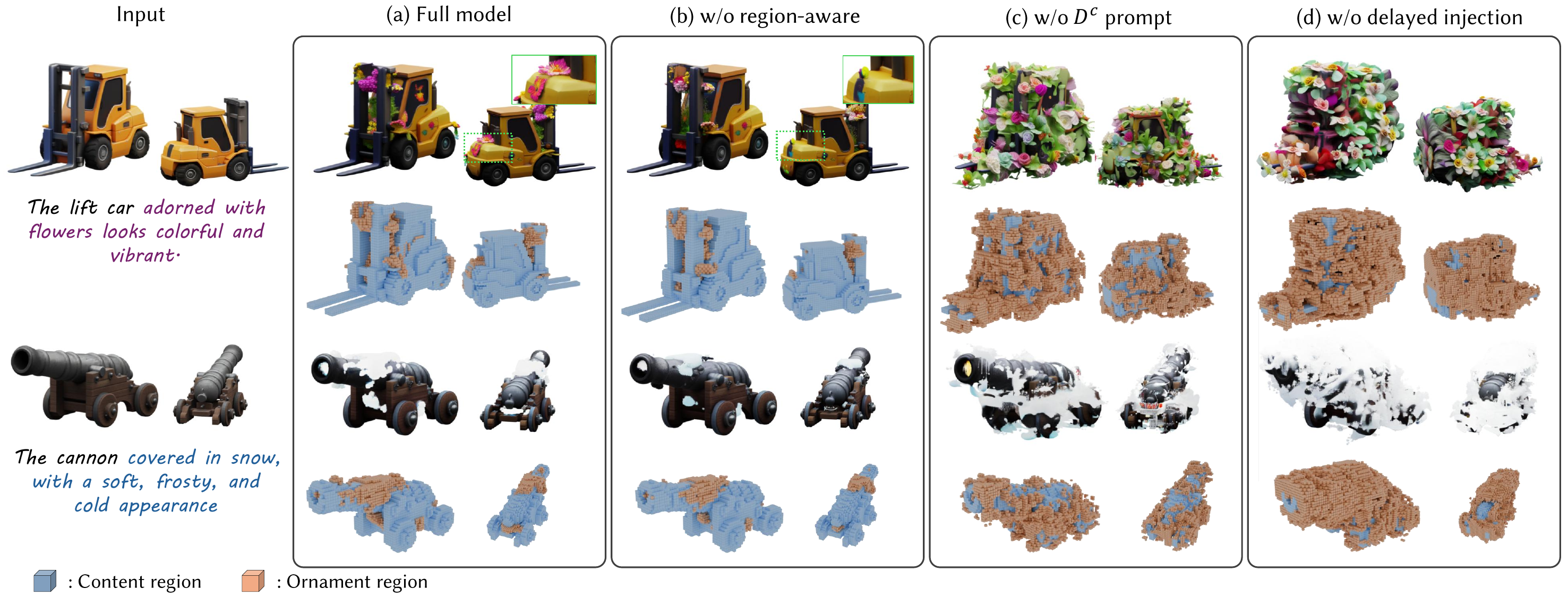}
    \caption{\textbf{Component ablation} for the voxel-level editing stage. The final edited results are presented together with voxel-level visualizations, in which blue cubes represent the content region, and the orange ones represent the additional ornamentation region.}
    \label{fig: abl_all}
\end{figure*}

\noindent \textbf{Semantic Control.} 
OrnaStyler further supports semantic refinement of ornamental content. Specifically, by replacing the original style-related textual guidance $D^s$ with more detailed descriptions during updating with Eq. \ref{eq:update_latent_harmoni} for feature inpainting in the latent-level style editing stage, users can alter the semantic identity of the ornaments while preserving their overall spatial configuration. As illustrated in Fig.~\ref{fig: flowers}, a generic flower arrangement can be semantically refined into different identities, such as \textit{Marigold}, while maintaining the same spatial layout. In addition, more fine-grained textual specifications allow further semantic control, e.g., changing \textit{Rose} into \textit{White rose} or \textit{Purple rose} (red dotted area). This confirms the flexibility of our method in enabling desired semantic-level detail control at both ornament identities and attributes by simply preparing different $D^s$ for feature harmonization.

\subsection{Analysis}
\noindent \textbf{Component Ablation.}
The voxel-level ornament synthesis stage in our method includes the following three key components: (a) region-aware guidance, (b) $D^c$ guidance, and (c) delayed injection. We here respectively study the effectiveness. 
Specifically, for (a), we remove the occupancy mask $M^o$ and guide the entire regions with only $D^e$ in Eq.~\ref{eq:tiled_v^s}; for (b), we exclude content description $D^c$ in the spatial region by substituting the additional style description $D^s$ for edited description $D^e$ in Eq.~\ref{eq:tiled_v^s}; and for (c) we assess the effectiveness of delayed injection by replacing the initial noise $\mathbf{S}^c_{t_{\alpha N}}$ with fully inverted noise $\mathbf{S}^c_{t_N}$ in Eq.~\ref{eq: S^s_t_i-1} (i.e., setting $\alpha$ to 1).

The visualization comparison is given in Fig.~\ref{fig: abl_all}.
Although all ablation scenarios introduce ornamental structures, they exhibit less natural results compared to the full model. Without region-aware guidance (Fig.~\ref{fig: abl_all}(b)), the generated ornamentation lacks alignment with the underlying content surface, leading to spatial incoherence. This also restricts the effective allocation of style voxels, as content-aware constraints are removed. Also, when removing content guidance (Fig.~\ref{fig: abl_all}(c)), excessive ornamentation is introduced, which damages the original structure and degrades content identity. This is because the inversion latent, although derived from $D^c$, is no longer explicitly constrained during generation, which causes the model to over-emphasize stylistic patterns.
A similar effect is observed when disabling delayed injection (Fig.~\ref{fig: abl_all}(d)). Here, a full inversion to $\mathbf{S}^c_{t_N}$ weakens structural priors, resulting in uncontrolled ornament propagation. Moreover, the fact that shape deviations from the original geometry caused by erroneous reconstruction are interpreted as stylistic structures further amplifies over-generation. Overall, our full model achieves the best content-style harmonization with the most natural visual effect. 

\begin{figure}[t]
    \centering
    \includegraphics[width=\linewidth]{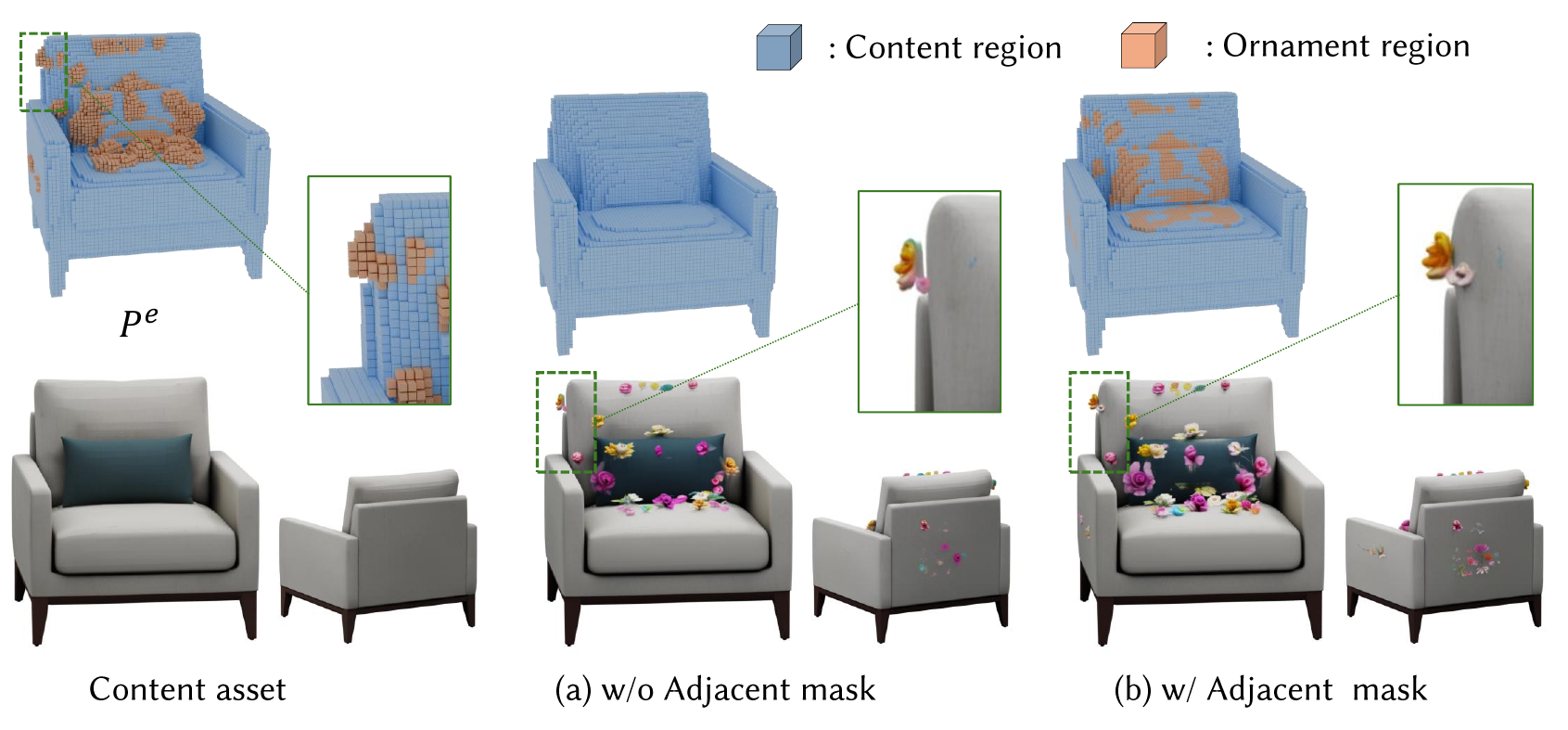}
    \caption{\textbf{Influence of adjacency-aware masking.} }
    \label{fig: abl_mask}
\end{figure}

\begin{figure}[t]
    \centering
    \includegraphics[width=\linewidth]{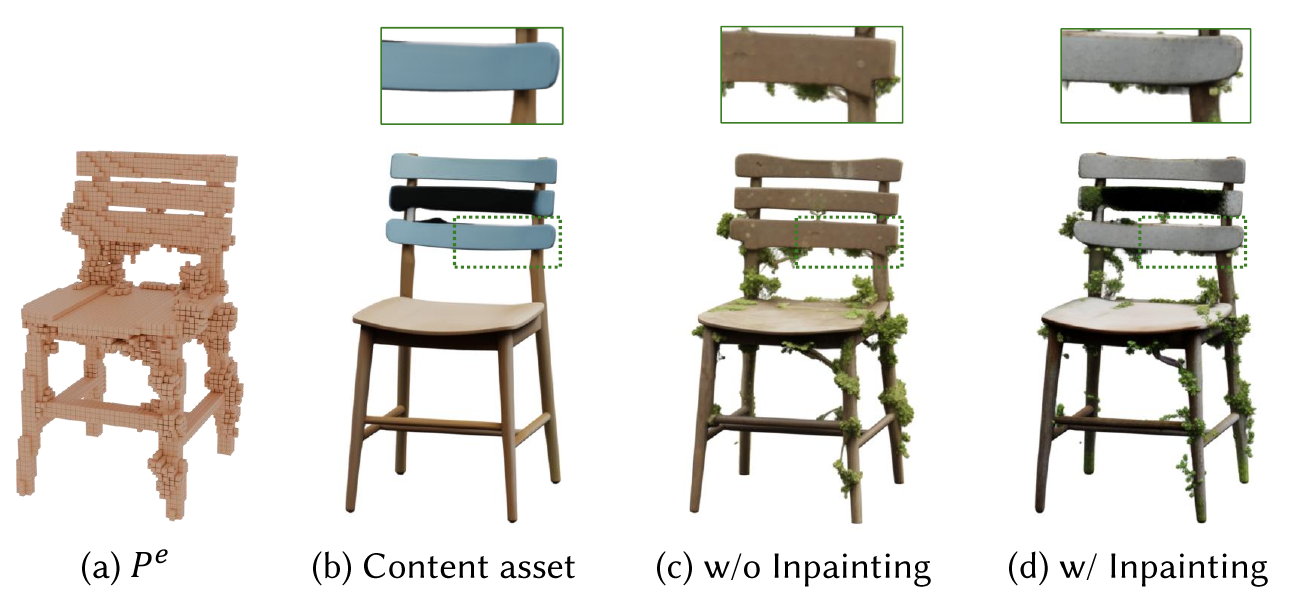}
    \caption{\textbf{Influence of inpainting-based feature integration.} }
    \label{fig: abl_pe}
\end{figure}

\noindent \textbf{Influence of Adjacency-Aware Masking.}
To mitigate unintended distortions near the boundaries between content and ornament regions, OrnaStyler incorporates a spatial mask that explicitly captures adjacency between ornaments and content. To evaluate the effect, we ablate it (i.e., setting the 1-valued entry for $M^l$ in Eq. \ref{eq:inpaint_feature} to the active voxel region of $\mathbf{P}^s$ ). As shown in Fig.~\ref{fig: abl_mask}, w/o adjacent mask (b) exhibits discontinuous geometry and broken structural connections between the content and the ornamentation, as the flower together with a small part of the sofa texture appears isolated from the sofa itself. This is because pure generation without adjacency modeling 
fails to capture the spatial coherence between the content and the ornaments during latent feature generation. 

\noindent \textbf{Influence of Inpainting-Based Feature Integration.}
Our method incorporates an inpainting-based strategy to construct the final latent features.
To evaluate its effectiveness, we design a baseline, where the structured latent $\mathbf{F}^e$ is generated from scratch using only the stylized description $D^e$ via $\mathcal{G}_L$ without applying the inpainting mechanism. The comparison is presented in Fig.~\ref{fig: abl_pe}. 
The baseline without inpainting (c) exhibits noticeable degradation in fine-grained content details, including both geometry and texture. 
This is because the absence of spatial and semantic constraints causes the model to treat ornament regions and content regions uniformly, leading to the misinterpretation of the added ornament structures as part of the original object (e.g., the highlighted area).
In contrast, our inpainting-based formulation enforces a clear separation between ornament and content, allowing each component to be generated under appropriate constraints before integration. The above result validates the effectiveness of inpainting-based feature integration in harmonizing both content and style characteristics during editing.

\begin{figure}[t]
    \centering
    \includegraphics[width=\linewidth]{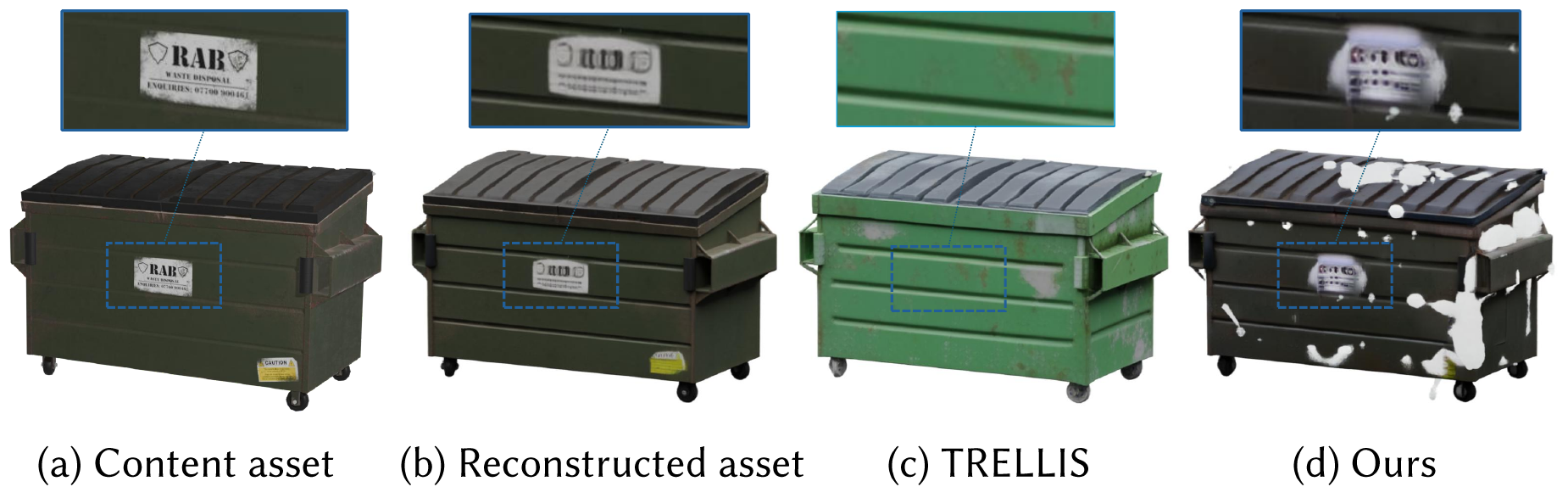}
    \caption{\textbf{Failure cases}
     in preserving detailed designs. }
    \label{fig: limitation}
\end{figure}

\section{Limitation}
Our method builds upon TRELLIS as the generative backbone, and is therefore inherently limited by its representational capacity for 3D shapes. In particular, we observe that detailed designs, such as texts, are often blurred or poorly preserved (Fig.~\ref{fig: limitation}(c, box)). This is mainly because the pre-trained flow matching model in TRELLIS struggles to reconstruct high-frequency textures (Fig.~\ref{fig: limitation}(b)), which limits the fidelity of fine details in our results. Nevertheless, OrnaStyler still improves over TRELLIS in both style reflection and content consistency (Fig.~\ref{fig: limitation}(d)). This limitation could potentially be mitigated by adopting more expressive generative backbones that handle higher-resolution or multi-scale latent representations, which we would like to explore in the future. Also, our formulation focuses on additive ornament editing and does not support subtractive operations that require removing the original content geometry (e.g., erosion). In addition, our method can be advanced to adapt intra-style masking strategies to realize multi-style complex decorative editing,  which serves as another interesting future direction.

\section{Conclusion}
We have presented OrnaStyler, a zero-shot text-guided 3D asset editing framework that specifically models the spatial configuration of ornaments for stylization. It leverages flow inversion to effectively trace the seed latents at both the voxel and appearance level to prevent structural corruption while pursuing style-faithful and content-preserving stylization. Moreover, we introduce an adjacency-aware spatial mask to ensure visual consistency around content-ornament boundaries for further content-style harmonization. Extensive experiments demonstrate that OrnaStyler achieves state-of-the-art performance in ornament-aware 3D shape editing compared with prior methods, both qualitatively and quantitatively.

{
    \small
    \bibliographystyle{ieeenat_fullname}
    \bibliography{main}
}

\clearpage

\appendix
\section{Setting and evaluation metrics}\label{sec: appendix_qualitative}
\noindent \textbf{Rendering Setting}. 
For quantitative evaluation, each generated sample is rendered from 12 fixed camera poses.
The camera is placed along a circular trajectory centered at the object origin with a fixed elevation angle of 30°.
The azimuth angles are uniformly sampled at 30° intervals over the range [0°, 360°).
All cameras are oriented toward the object center with a fixed direction from above.

\noindent \textbf{Evaluation Metrics}.\label{appn: metrics}
To evaluate the similarity between content and edited results, we adopt standard metrics widely used in 3D generation and editing literature \cite{liu2024text,qu2025stylesculptor,xiang2025structured}:

\begin{itemize}
  \item \textbf{Structure SIMilarity (SSIM)}: measures the structural correspondence between the rendered images of the original content and the edited 3D asset by comparing luminance, contrast, and structural information.
  
  \item \textbf{Peak Signal-to-Noise Ratio (PSNR)}: evaluates pixel-wise fidelity by measuring the ratio between the maximum signal and reconstruction noise.
  
  \item \textbf{Learned Perceptual Image Patch Similarity (LPIPS)}: computes perceptual distance using deep features to provide a similarity measure aligned with human perception.
  
  \item \textbf{Contrastive Language-Image Pre-Training (CLIP)}: measures cosine similarity between image embeddings in the CLIP feature space to reflect semantic alignment.
  
  \item \textbf{Fréchet Inception Distance (FID)}: evaluates distributional similarity between sets of rendered images in feature space to capture visual quality.
  
  \item  \textbf{Intersection-over-Union (IoU)}: quantifies volumetric consistency by computing the overlap between occupied voxel regions of the content and edited shapes.
  
  \item  \textbf{Chamfer Distance (CD)}: computes the bidirectional distance between point sets sampled from voxel centroids to evaluate geometric consistency in 3D space.

\end{itemize}

\section{Efficiency analysis.}\label{appn: efficiency}
To evaluate the computational efficiency of text-driven 3D style editing, we report the average runtime of different methods in Tab.~\ref{tb:score_time} on both datasets. As OrnaStyler is built upon the TRELLIS framework, it introduces additional computational overhead to handle ornament-aware processing, including voxel augmentation and feature harmonization. Despite this, our method achieves the second-best runtime among all compared approaches. This result indicates that OrnaStyler maintains a favorable trade-off between efficiency and editing quality.

\section{Parameter sensitivity}\label{appn: parameter}
We investigate the delayed injection rate $\alpha$ on four representative assets with contrasting structures: thin structures: the helicopter (in Fig.~\ref{fig: appendix_1}) and desk lamp (in Fig.~\ref{fig: appendix_2}), and dense structures: the cart (in Fig.~\ref{fig: pipeline}) and sofa (in Fig.~\ref{fig: abl_mask}). Specifically, we vary $\alpha$ within $[0.5,1.0]$, and report (1) CLIP similarity for content consistency; (2) the Surface Coverage Ratio (SCR) for ornament coverage, defined as
$\mathrm{SCR}=|\{p \in P^c \mid p \text{ is face-adjacent to a generated ornament voxel}\}|/|P^c|$.
As shown in Tab.~\ref{tab:alpha_sensitivity}, both structure types exhibit the same trade-off: a smaller $\alpha$ produces limited ornamentation, whereas a larger $\alpha$ increases ornament coverage at the cost of content consistency.
We therefore apply $\alpha=0.8$ to provide balanced results for diverse structures.

\section{Prompt types}\label{appn:prompt}
\begin{figure}
    \centering
    \includegraphics[width=\linewidth]{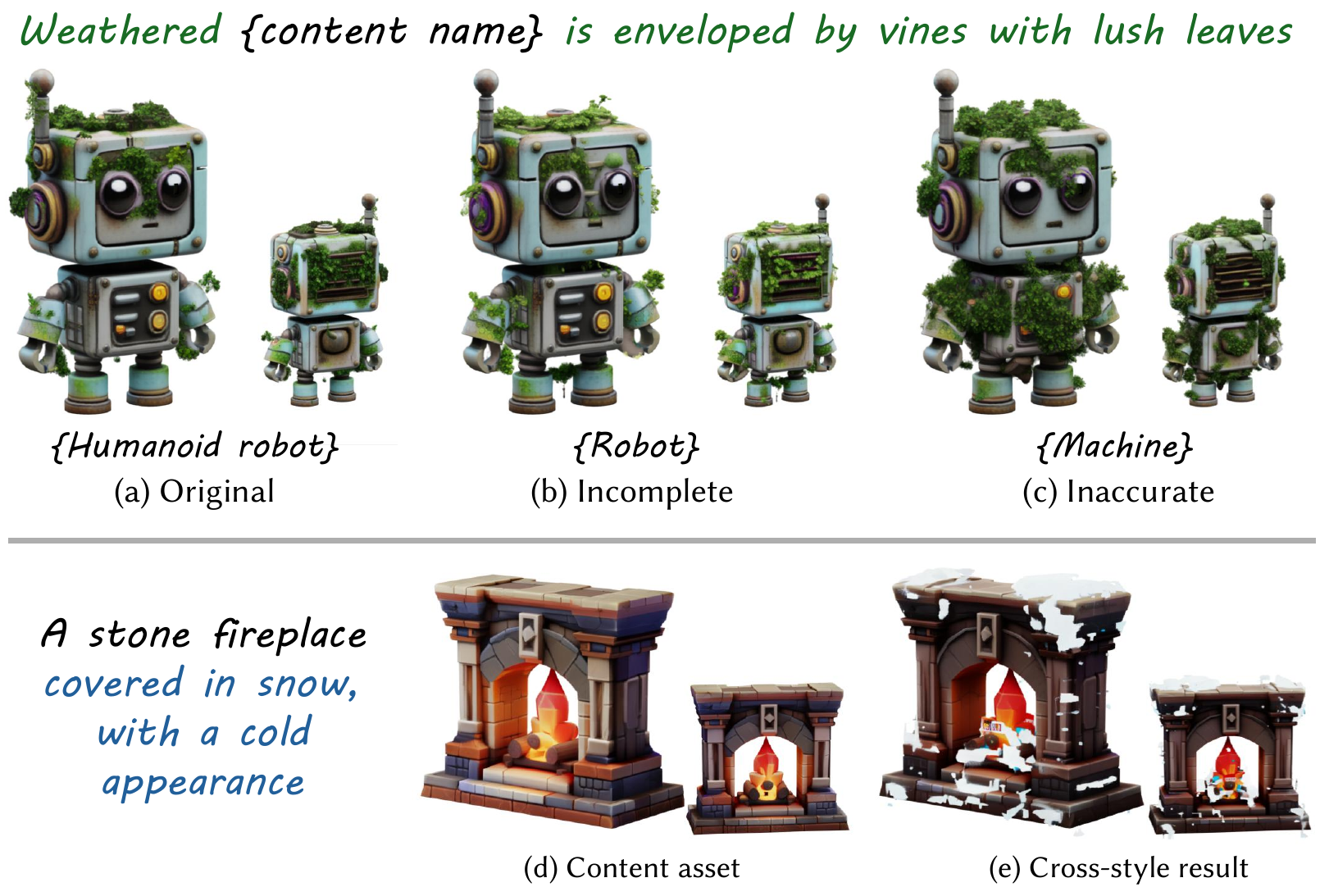}
    \caption{
    \textbf{Robustness to different prompt types.} 
    Top: stylization results with (a) original, (b) incomplete, and (c) inaccurate content prompts.
    Bottom: cross-semantic stylization (e) of a fire-themed asset (d) with the \textit{snow} style.
    }
    \label{fig:prompt_types}
\end{figure}

We here discuss cases in which the given prompts are incomplete/inaccurate or with semantic conflicts. As depicted in Fig. \ref{fig:prompt_types} (top row), when replacing “Humanoid robot” with the less specific “Robot” and the inaccurate “Machine”, the resulting stylizations still remain visually consistent, indicating that OrnaStyler is reasonably robust to moderate prompt imprecision, as the source geometry and appearance are also anchored by inversion of the input asset. We further evaluate semantically contrasting content and style conditions by applying the \textit{snow} style to a fire-themed 3D asset in Fig.~\ref{fig:prompt_types} (bottom row). Despite the semantic conflict of fire and snow, our method still integrates 
the contrasting geometries coherently, which demonstrates that  OrnaStyler also realizes cross-semantic stylization. Nevertheless, OrnaStyler does not model physical interactions or temporal effects, and therefore cannot simulate causal responses (e.g., ice melting).

\begin{table}[t]
    \centering
    \renewcommand{\arraystretch}{1.0}
    \caption{\textbf{Timing evaluations (in seconds) for different techniques.} We calculate the runtime on both TRELLIS-Generated and Sketchfab data. }
    \resizebox{\linewidth}{!}{
        \begin{tabular}{lcc}
            \toprule
                                                    \textbf{Method} &{\textbf{TRELLIS-Generated}} &{\textbf{Sketchfab}}\\  \hline 
            Text2Tex \textit{\fontsize{7}{10}\selectfont[ICCV'23]}        &   225.9 &  227.9  \\ 
            TEXTure \textit{\fontsize{7}{10}\selectfont[SIGGRAPH'23]}     &   485.2 &  481.1 \\ 
            TexPainter \textit{\fontsize{7}{10}\selectfont[SIGGRAPH'24]}    &   225.3 &  178.9  \\ 
            SyncMVD \textit{\fontsize{7}{10}\selectfont[SIGGRAPH'24]}       &   47.9 &  44.3 \\ 
            Paint3D \textit{\fontsize{7}{10}\selectfont[CVPR'24]}           &   29.4 &  31.0 \\   
            TRELLIS \textit{\fontsize{7}{10}\selectfont[CVPR'25]} + ICEdit \textit{\fontsize{7}{10}\selectfont[NeurIPS'25]}           &   43.1 &  47.2 \\ 
            TRELLIS \textit{\fontsize{7}{10}\selectfont[CVPR'25]} + Gemini        &   31.8 &  31.3 \\ 
            TRELLIS \textit{\fontsize{7}{10}\selectfont[CVPR'25]}      &   2.2 &  3.1 \\ 
            \textbf{OrnaStyler}                                   &   21.8 &  22.4  \\               
            
            \bottomrule
        \end{tabular}
    }
    \label{tb:score_time}
\end{table}

\begin{table}[t]
    \centering
    \renewcommand{\arraystretch}{1.0}
    \caption{\textbf{Effect of the delayed injection rate $\alpha$.}}
    \resizebox{0.6\linewidth}{!}{
        \begin{tabular}{c|cccc}
            \toprule
            \multicolumn{1}{c}{} 
            & \multicolumn{2}{c}{\textbf{Thin}}
            & \multicolumn{2}{c}{\textbf{Dense}} \\
            \cmidrule(lr){2-3} \cmidrule(lr){4-5}

            $\alpha$ & {CLIP} & {SCR} & {CLIP} & {SCR} \\
            \hline

            0.6 & 0.99 & 0.03 & 0.99 & 0.01 \\
            0.7 & 0.98 & 0.07 & 0.99 & 0.02 \\
            0.8 & 0.92 & 0.17 & 0.93 & 0.20 \\
            0.9 & 0.80 & 0.38 & 0.85 & 0.43 \\
            1.0 & 0.74 & 0.52 & 0.78 & 0.60 \\

            \bottomrule
        \end{tabular}} 
    \label{tab:alpha_sensitivity}
\end{table}

\section{User study} \label{appn: user_study}
\begin{figure}[t]
    \centering
    \includegraphics[width=1.0\linewidth]{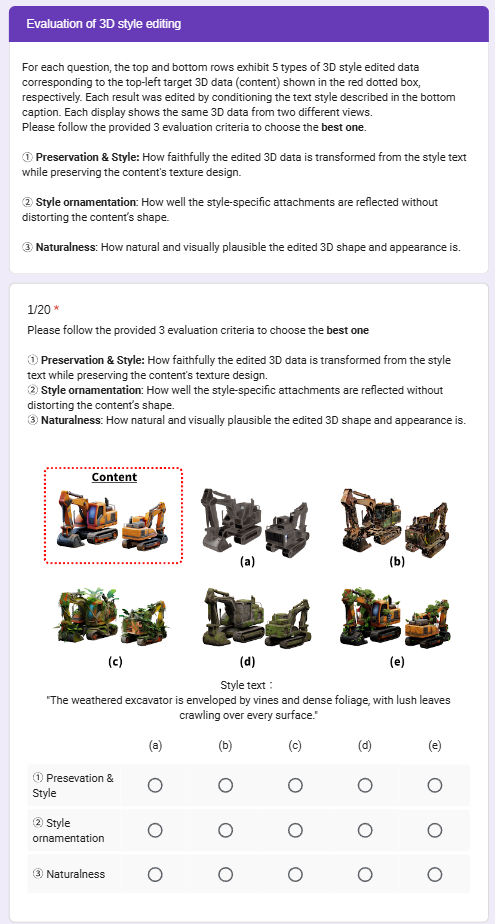}
    \caption{
    \textbf{User study interface} with the instructions for participants (top). }
    \label{fig: userstudy_format}
\end{figure}

We provide additional details of our user study. The study involves a total of 30 participants, including 14 internal participants with backgrounds in machine learning and 16 external participants without professional expertise. None of the participants received compensation. The user interface of our study is illustrated in Fig.~\ref{fig: userstudy_format}. Each study session takes approximately 20–25 minutes to complete, consisting of 20 questions with 3 evaluation criteria per question.

Before the evaluation, participants are given detailed explanations of each assessment criterion to ensure consistent understanding. For each question, we present the style-edited results generated by different methods in a randomized order to avoid bias. The compared methods include TEXTure \cite{richardson2023texture}, Paint3D \cite{zeng2024paint3d}, Gemini \cite{comanici2025gemini}, TRELLIS \cite{xiang2025structured}, and our method. Specifically, for each question, the rendered views (front and back) of the original content 3D asset are displayed in the upper-left corner. The editing results from different methods are shown under identical viewpoints to ensure fair comparison. The corresponding style text prompt is also provided for reference.

\section{Asset attribution} \label{appn: sketchfab}
We utilize 3D assets obtained from Sketchfab under the Creative Commons Attribution 4.0 International (CC BY 4.0) license. Each model from Sketchfab is attributed as follows:
\begin{itemize}
  \item "Asylum Bed" by FallenBranches
  \item "Tiny Tikes Kids Table" by bungleboy
  \item "Iron Rack Shelf for Home Storage" by jackk
  \item "Kids Balance Bike" by bagussper
  \item "Low Poly Viking Drinking horn" by LachieRobertson
  \item "Kids Trike" by bungleboy
  \item "Stylised PBR Sci-Fi Chair" by Goob
  \item "Pirate Cannon" by Maxwell McCurry
  \item "Table" by Anch0r
  \item "Desk lamp" by GokouFG
  \item "WUD TOOLBOX HP" by Geug
  \item "Adidas Bad Bunny Coffee Shoes Free" by camilooh
  \item "Dumpster" by TheMadraver
  \item "Headset" by Elite Big Speakerman
  \item "WOODEN DUCK TOY" by UJJWAL CHAUHAN
  \item "PPC200" by Interiors3D
  \item "Old wall telephone" by renaud lapierre
  \item "Old Sofa" by Amad Junaid
  \item "Old Rowboat" by TooManyDemons
  \item "QWTA00" by Interiors3D
\end{itemize}

\begin{algorithm}[t]
    \begin{algorithmic}[1]
        \Require{{Content asset $A^c$, Content text $D^c$, Style text $D^e$, Pre-trained TRELLIS model $(\mathcal{E},\mathcal{E}_S,\mathcal{D}_S$, velocity $\bm{v}^s$ predicted by $\mathcal{G}_S)$, Number of denoising iteration steps $N$, Delay rate $\alpha$, Guidance strength $w_s$}}
        \Ensure{{Stylized coordinates $\mathbf{P}^e$}}
                \State {$\mathbf{Z}^c \leftarrow{} \mathcal{E}(A^c)$      \Comment{$\mathbf{Z}^c = \{ \mathbf{P}^c, \mathbf{F}^c\}$}}
                \State{$\hat{\mathbf{S}}^c_{t_0} \leftarrow \mathcal{E}_S(\mathbf{P}^c)$           \Comment{$\hat{\bm{v}}_{t_0} \leftarrow \bm{v}^s( \hat{\mathbf{S}}^c_{t_0}, t_0, D^c )$} }
                \State{/* \textit{Inversion} */}
                \For{{$t_i = t_0, ....,t_{\alpha N}$ }} 
                    \State $\tilde{\mathbf{S}}^c_{t_{i+1}} \leftarrow \hat{\mathbf{S}}^c_{t_{i}} - (t_i-t_{i+1})\bm{v}^s_{t_{i}}, \bm{v}^s_{t_{i+1}} \leftarrow \bm{v}^s( \tilde{\mathbf{S}}^c_{t_{i+1}},t_{i+1}, D^c )$ 
                    \State {$\hat{\mathbf{S}}^c_{t_{i+1}} \leftarrow \hat{\mathbf{S}}^c_{t_i} - (t_i-t_{i+1})\bm{v}^s_{t_{i+1}}$ }
                \EndFor{}
                \State {$\mathbf{S}^s_{t_{\alpha N}} \leftarrow \hat{\mathbf{S}}^c_{t_{\alpha N}}$ }   
                \State{/* \textit{Generation} */}
                \State {Obtain $M^o$ from Eq. (5) in Section 4.1 \Comment{$M^o \in \{0, 1\}$} }
                \For{{$t_i = t_{\alpha N}, ....,t_0$ } }
                    \State {$\bm{v}^{cfg}_{t_{i}} \leftarrow  w_s(\bm{v}^s(\mathbf{S}^s_{t_i}, t_i, D^e) -\bm{v}^s(\mathbf{S}^s_{t_i}, t_i, D^c)) $}
                    \State {$\bar{\mathbf{S}}^s_{t_{i}} \leftarrow \mathbf{S}^s_{t_{i}} + (t_{i-1} - t_{i})(1 - M^o) \odot \bm{v}^{cfg}_{{t_i}}   $}
                    \State {$\tilde{\bm{v}}^{{s}}_{t_i} \leftarrow M^o \odot \bm{v}^s(\mathbf{S}^s_{t_i}, t_i, D^c) + (1 - M^o) \odot \bm{v}^s(\mathbf{S}^s_{t_i}, t_i, D^e)$}
                    \State {$\mathbf{S}^s_{t_{i-1}} \leftarrow   \bar{\mathbf{S}}^s_{t_{i}} + (t_{i-1} - t_{i})\tilde{\bm{v}}^{{s}}_{t_i}$}
                \EndFor{}
                \State{$\mathbf{P}^e\leftarrow \mathcal{D}_S(\mathbf{S}^s_{t_0}) \cup \mathbf{P}^c $ }
                \State {\textbf{return} $\mathbf{P}^e$}
    \end{algorithmic}
    \caption{Voxel-level Ornament Synthesis}
    \label{alg: voxel-level}
\end{algorithm}

\begin{algorithm}[t]
    \begin{algorithmic}[1]
        \Require{{Content asset $A^c$, Style coordinates $\mathbf{P}^e$, Content text $D^c$, Edited text $D^e$, Style text $D^{s}$, Pre-trained TRELLIS model $(\mathcal{E},\mathcal{D}$, velocity $\bm{v}^f$ predicted by $\mathcal{G}_L)$, Number of denoising iteration steps $N$}}
        \Ensure{{Stylized asset $A^e$}}
                    \State {$\mathbf{P}^c, \mathbf{F}^c \leftarrow{} \mathcal{E}(A^c)$ }\Comment{$\hat{\mathbf{F}}_{t_0}^c \leftarrow{} \mathbf{F}^c, \bm{v}^f_{t_{0}} \leftarrow \bm{v}^f( \hat{\mathbf{F}}^c_{t_{0}},t_{0}, D^c )$}
                    \State{/* \textit{Content latent feature editing} */}
                        \For{{$t_i = t_0, ....,t_{N-1}$ }}                          \Comment{$ t_0 = 0, t_N = 1$}      
                            \State $\tilde{\mathbf{F}}^c_{t_{i+1}} \leftarrow \hat{\mathbf{F}}^c_{t_{i}} - (t_i-t_{i+1})\bm{v}^f_{t_i}, \bm{v}^f_{t_{i+1}} \leftarrow \bm{v}^f( \tilde{\mathbf{F}}^c_{t_{i+1}},t_{i+1}, D^c )$
                            \State {$\hat{\mathbf{F}}^c_{t_{i+1}} \leftarrow \hat{\mathbf{F}}^c_{t_i} - (t_i-t_{i+1})\bm{v}^f_{t_{i+1}}$ }
                        \EndFor{}
                        \State {$\mathbf{F}^{cs}_{t_{0}} \leftarrow \mathcal{G}_L(\hat{\mathbf{F}}^c_{t_{N}} | \mathbf{P}^c,  D^e) $  \Comment{$\mathbf{F}^{cs} = \{ f_{j}\}^{L}_{j = 1}$} }

                    \State{/* \textit{Adjacency modeling} */}                   
                        \State {$\mathbf{P}^d \leftarrow \mathbf{P}^s\leftarrow \mathbf{P}^e \setminus \mathbf{P}^c$ }
                        \State {$\mathbf{P}^d \leftarrow  \text{Dil}(\mathbf{P}^d)$}
                        \State {$M^l \leftarrow (\mathbf{P}^d \cap \mathbf{P}^c) \cup \mathbf{P}^s$ \Comment{$M^l \in \{0, 1\}$}}

                    \State{/* \textit{Ornament inpainting} */}
                        \State {$\mathbf{F}^e_{t_N} \sim \mathcal{N}(\mathbf{0}, \mathbf{I})$ \Comment{Sampling} }
                        \For{{$t_i = t_{N-1}, ....,t_0$ }} 
                            \State {$\tilde{\mathbf{F}}^e_{t_{i-1}} \leftarrow   \mathbf{F}^e_{t_{i}} + (t_{i-1} - t_{i})\bm{v}^f(\mathbf{F}^e_{t_i}, t_i , \mathbf{P}^e, D^s)$} 
                            
                            \State{$\mathbf{F}^e_{t_{i-1}} \leftarrow M^l \odot \tilde{\mathbf{F}}^e_{t_{i-1}} + (1 - M^l) \odot \psi_{t_{i-1}}(\mathbf{F}^{cs}) $  }
                        \EndFor{}
                        \State{$A^e \leftarrow \mathcal{D}(\mathbf{Z}^e)$ \Comment{$\mathbf{Z}^e =\{ \mathbf{F}^e_{t_0}, \mathbf{P}^e \}$}}
                    \State {\textbf{return} $A^e$}
    \end{algorithmic}
    \caption{Latent-level Style Editing}
    \label{alg:inference2}
\end{algorithm}

\section{Algorithm} \label{appn: algorithm}
Our overall algorithm for both stages of our method is given in Alg. \ref{alg: voxel-level} and Alg. \ref{alg:inference2}, respectively.

\section{More results}  
\begin{figure*}[t]
    \centering
    \includegraphics[width=\linewidth]{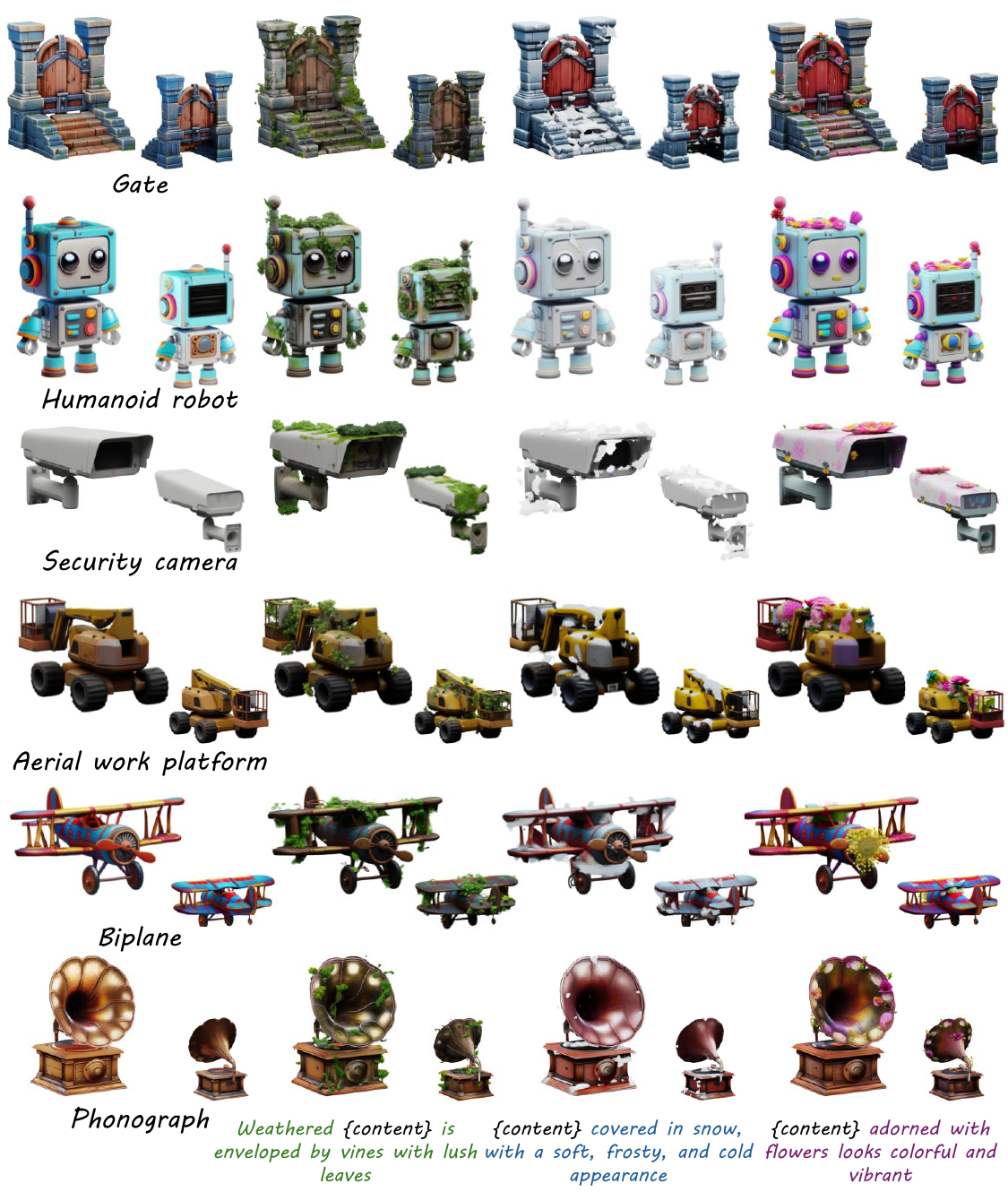}
    \caption{\textbf{Qualitative  results} of asset editing by OrnaStyler.  \textcolor{OliveGreen}{Colored}  \textcolor{NavyBlue}{textual}  \textcolor{RedViolet}{prompts} denote the style texts for editing the content asset generated with the black texts. }
    \label{fig: qualitative}
\end{figure*}

\begin{figure*}[t]
    \centering
    \includegraphics[width=\linewidth]{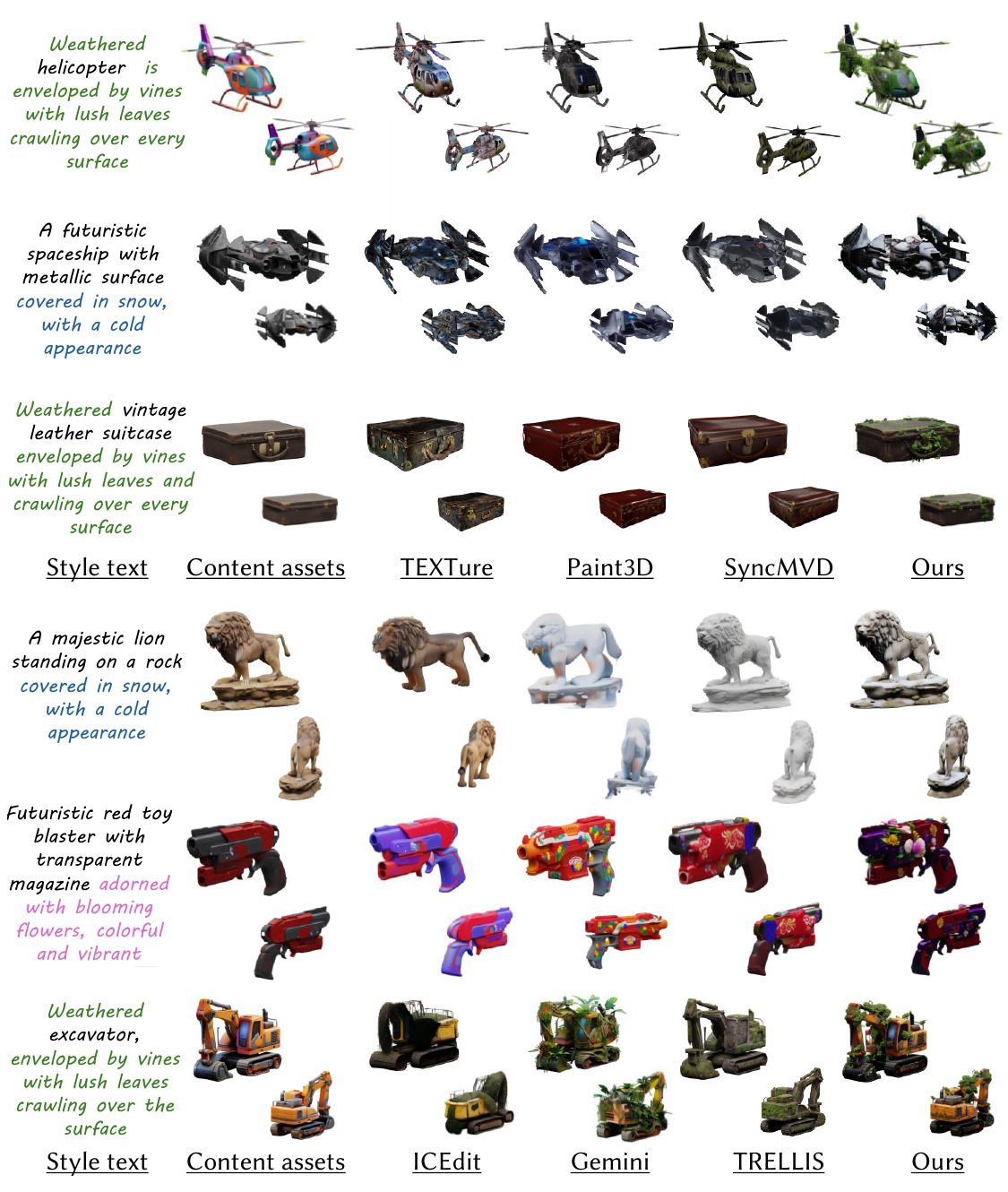}
    \caption{
    \textbf{Visualization of comparisons} against prior methods on Trellis-Generated data.}
    \label{fig: appendix_1}
\end{figure*}

\begin{figure*}[t]
    \centering
    \includegraphics[width=\linewidth]{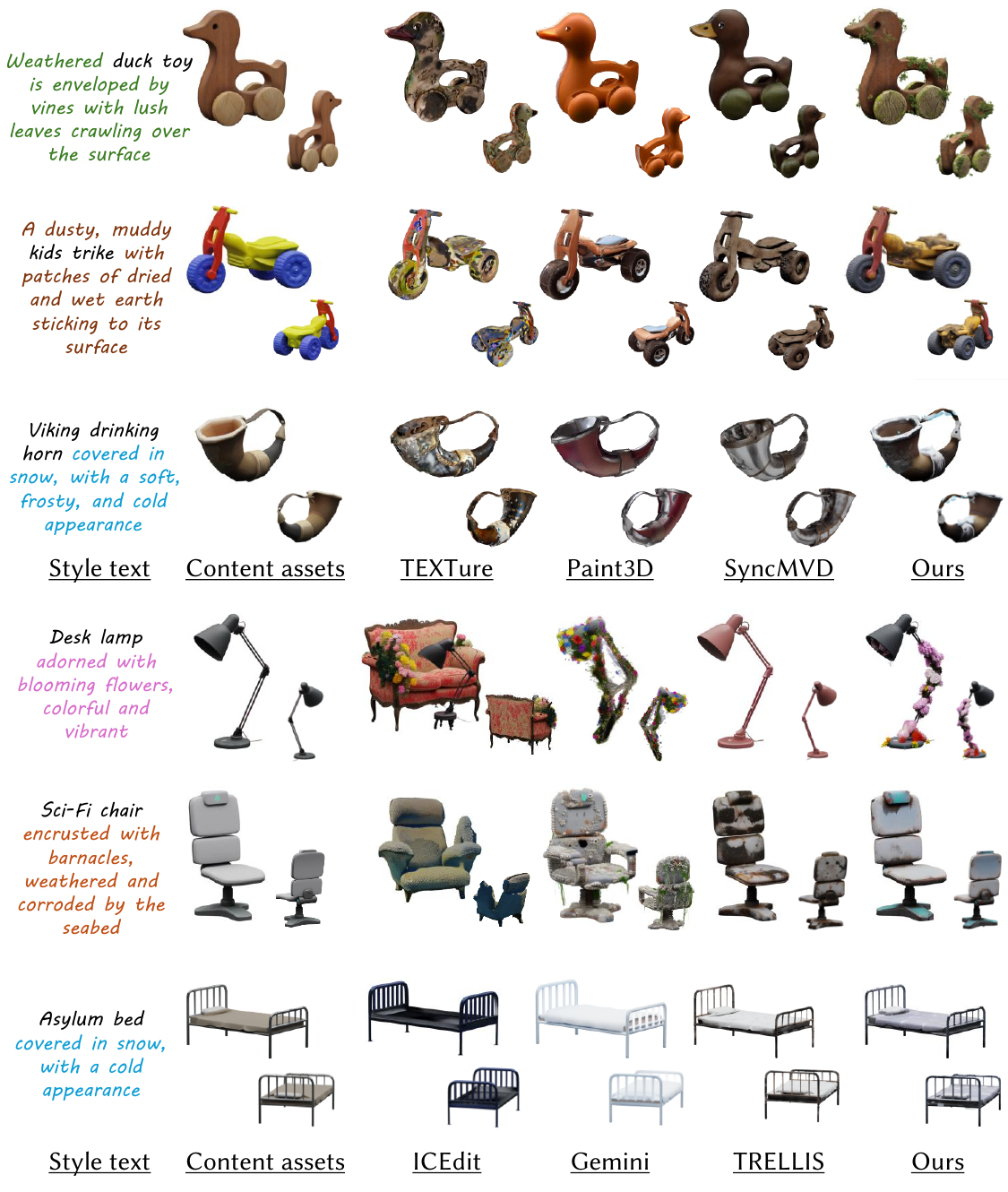}
    \caption{
    \textbf{Visualization of comparisons}  against prior methods on Sketchfab data.}
    \label{fig: appendix_2}
\end{figure*}

We show more style editing results of our method in Fig. \ref{fig: qualitative} and visual comparisons in Figs. \ref{fig: appendix_1} and  \ref{fig: appendix_2}. 

\end{document}